# A novel multi-agent dynamic portfolio optimization learning system based on hierarchical deep reinforcement learning

Ruoyu Sun[a], Yue Xi[b], Angelos Stefanidis[c], Zhengyong Jiang[d,*], Jionglong Su[e,*]

[a] School of AI and Advanced Computing, XJTLU Entrepreneur College (Taicang), Xi'an Jiaotong-Liverpool University, Suzhou, 215123, Jiangsu, China. Ruoyu.Sun19@student.xjtlu.edu.cn, https://orcid.org/0009-0002-6052-0051

[b] Department of Educational Studies, School of Academy of Future Education, Xi'an Jiaotong-Liverpool University, Suzhou, 215123, Jiangsu, China. Yue.Xi22@student.xjtlu.edu.cn, https://orcid.org/0009-0003-5373-6126

[c] School of AI and Advanced Computing, XJTLU Entrepreneur College (Taicang), Xi'an Jiaotong-Liverpool University, Suzhou, 215123, Jiangsu, China. Angelos.Stefanidis@xjtlu.edu.cn, https://orcid.org/0000-0002-4703-8765

[d] School of AI and Advanced Computing, XJTLU Entrepreneur College (Taicang), Xi'an Jiaotong-Liverpool University, Suzhou, 215123, Jiangsu, China. Zhengyong.Jiang02@xjtlu.edu.cn, https://orcid.org/0000-0001-8873-4073

[e] School of AI and Advanced Computing, XJTLU Entrepreneur College (Taicang), Xi'an Jiaotong-Liverpool University, Suzhou, 215123, Jiangsu, China. Jionglong.Su@xjtlu.edu.cn, https://orcid.org/0000-0001-5360-6493

* Corresponding author.

**Abstract**

Deep Reinforcement Learning (DRL) has been extensively used to address portfolio optimization problems. The DRL agents acquire knowledge and make decisions through unsupervised interactions with their environment without requiring explicit knowledge of the joint dynamics of portfolio assets. Among these DRL algorithms, the combination of actor-critic algorithms and deep function approximators is the most widely used DRL algorithm. Here, we find that training the DRL agent using the actor-critic algorithm and deep function approximators may lead to scenarios where the improvement in the DRL agent's risk-adjusted profitability is not significant. We propose that such situations primarily arise from the following two problems: sparsity in positive reward and the curse of dimensionality. These limitations prevent DRL agents from comprehensively learning asset price change patterns in the training environment. As a result, the DRL agents cannot explore the dynamic portfolio optimization policy to improve the risk-adjusted profitability in the training process. To address these problems, we propose a novel multi-agent Hierarchical Deep Reinforcement Learning (HDRL) algorithmic framework in this research. Under this framework, the agents work together as an learning system for portfolio optimization. Specifically, by designing an auxiliary agent that works together with the executive agent for optimal policy exploration, the learning system can focus on exploring the policy with higher risk-adjusted return in the action space with positive return and low variance. In this way, we can overcome the issue of the curse of dimensionality and improve the training efficiency in the positive reward sparse environment. The performance of the proposed HDRL algorithm is evaluated using a portfolio of 29 stocks from the Dow Jones index in four different experiments. During training, the risk-adjusted profitability of the DRL agent in the training environment is significantly improved. Hence, we can prove that the strategies executed by our learning system in out-sample experiments originate from the DRL agents' comprehensive learning of asset price change patterns in the training environment. Furthermore, each back-test experiment compares the proposed learning system to sixteen traditional strategies and ten strategies based on machine learning algorithms in the performance of profitability and risk control ability. The empirical results in the four evaluation experiments demonstrate the efficacy of our learning system, which outperforms all other strategies by at least 6.3% in terms of return per unit risk. Moreover, our proposed HDRL algorithm framework also outperforms individual DRL agents in the ablation study framework by a margin of at least 9.7% in terms of return per unit risk.

# 1. Introduction

In portfolio optimization, investors allocate funds among assets within a portfolio based on their specific preferences for expected returns and risk. The objective is to maximize returns while controlling risk during each trading period by allocating funds across a range of assets. Investors express their preferences for expected returns and risk by constructing a specific objective function during the portfolio optimization process [1]. Typically, the objective function is defined as the expected return of the portfolio calibrated against its risk [2]. Traditional portfolio optimization strategies can be divided into two categories, i.e., strategies based on Markowitz's mean-variance theory [1-3] and strategies based on the Capital Growth Theory [4, 5]. The strategies belonging to the Capital Growth Theory can be classified into five types. They are respectively termed "Benchmarks", "Follow the winner", "Follow the loser", "Pattern-Matching Approaches", and "Meta-Learning Algorithms" [6].

With an increase in financial market data, as well as the complexity of influencing factors in the financial market, financial analysis has become increasingly challenging. However, with the advancement of Artificial Intelligence (AI), novel perspectives and solutions are proposed to address the challenges in finance. AI algorithms can enhance risk assessment and hedging [7], predict stock market trends [8-12], and realize statistical arbitrage [13]. The utilization of AI techniques in stock analysis and prediction has gained considerable traction due to its ability to handle large volumes of data as well as extract meaningful patterns and insights. AI is extensively used in the financial market for prediction and portfolio optimization [14, 15]. In portfolio optimization, AI is developed and explored as a method to address the challenges posed by the high volatility of financial assets [16]. Furthermore, the use of AI in portfolio optimization serves as an effective means to mitigate behavioral economic biases [17].

As an AI technique, Deep Reinforcement Learning (DRL) agents acquire knowledge and make decisions through unsupervised interactions with their environment. Since model-free DRL algorithms do not require explicit knowledge of the joint dynamics of portfolio assets, they find extensive use in learning systems such as that for portfolio optimization [18]. To effectively extract the information from the environment for portfolio optimization, various DRL algorithms train the DRL agent so that portfolio weights may be determined by the trained neural network. The algorithms for training DRL agents can be categorized into paradigms: policy-based algorithm [19-34] and value-based algorithm [35-41]. One major limitation of the value-based algorithms is that they can only handle discrete action spaces. Hence, when training an agent using value-based algorithms, it is necessary to discretize the action space. For a fine-grained discretization of the action space, the presence of a large number of assets in the investment portfolio leads to a significant increase in the number of actions that need to be explored. This can result in computational prohibitiveness when exploring the optimal policy and training DQN-like networks [42]. Moreover, the simplistic discretization of action spaces disregards valuable information pertaining to the action domain, potentially compromising the ability to effectively address various problem scenarios [42]. Hence, when managing investment portfolios with a large number of assets, employing policy-based algorithms is the preferred choice [43]. Within the policy-based approaches, the combination of actor-critic algorithms and deep function approximators (e.g., DDPG [42], SAC [44], PPO [45]) is widely used to train the DRL agent for portfolio optimization. This is primarily due to their ability to handle high-dimensional continuous spaces while leveraging Q-learning algorithms [42, 44, 45].

In existing research, a notable phenomenon is observed where scholars primarily focus on evaluating the performance of DRL agents using out-of-sample data while often neglecting to analyze the agents' improvements in profitability and risk control during the training process. As a result, such empirical results in the out-of-sample data cannot demonstrate that the strategies employed in back-testing experiments are derived from the DRL agent's thorough exploration of optimal strategies within the training environment. To address this issue, we develop a tracking module designed to monitor the trajectories of indicators that reflect improvements in the DRL agent's risk-adjusted profitability within the training environment. Additionally, this module tracks changes in the values of the objective functions utilized during the training of the DRL agent. Throughout the tracking process, we identify that training the DRL agent using the actor-critic algorithm and deep function approximators may lead to scenarios where the improvement in the agent's risk-adjusted profitability is not significant. This suggests that, when employing the actor-critic algorithm alongside deep function approximators for portfolio optimization over consecutive trading periods, the DRL agent faces challenges in improving the risk-adjusted return obtained in the training environment. Borrowing from the ideas of Aremu et al. [46] and Ocana [47], we argue that such situations primarily arise from the following problems:

- **Sparsity in positive reward.** To avoid excessive concentration of investment capital in a few

portfolio assets judged to yield high returns, we consider a balance between returns and risks when formulating the reward function in the environment. To realize the trade-off between the expected return and risk of the portfolio, the training objective function of the DRL agents for portfolio optimization is constructed in the form of the return modified by the assumption of risk. However, this results in a sparsity of positive rewards, making it difficult for the DRL agents to learn optimal policies in portfolio management.

- **Curse of dimensionality.** During the DRL agent training, the curse of dimensionality [48] refers to the exponential growth of states and actions when exploring optimal policy in high-dimensional spaces [49]. As a result, DRL agents suffer from poor sample efficiency and poor scalability [50]. In portfolio optimization, the growth in the number of assets in the portfolio and the market's allowance of short selling both can trigger the curse of dimensionality.

These problems prevent the critic network from accurately estimating the value of each action. To overcome these, we propose a more effective way to train the DRL agent, i.e., an learning system based on the multi-agent Hierarchical Deep Reinforcement Learning (HDRL) algorithm [51] for portfolio optimization in consecutive trading periods. Hierarchical reinforcement learning involves dividing the target portfolio optimization problem into a hierarchy of subproblems. By doing so, HDRL strengthens the learning process of individual abstractions within the hierarchy [52]. Moreover, it seeks to enable agents to solve complex problems by decomposing them into subproblems and learning high-level policies to guide the decision-making process [53].

The novelty of our work is as follows:

1) To the best of our knowledge, we are the first to highlight the importance of tracking the improvement in the DRL agent's profitability and risk-adjusted profitability in the training process. Furthermore, based on the ideas of Aremu et al. [46] and Ocana [47], we are the first to argue that, in the context of challenges associated with training a DRL agent using the actor-critic algorithm for dynamic portfolio optimization, the primary difficulties stem from the scarcity of positive rewards and the curse of dimensionality within the training environment. Based on this idea, we design an auxiliary DRL agent to carry out the auxiliary task that assists the DRL agent based on the actor-critic algorithm in the exploration of the optimal policy. In this way, the DRL agent based on the actor-critic algorithm can explore the optimal policy in the action space with positive return and low variance. Hence, the computational cost in the optimal policy exploration can be reduced.

2) To address the issues of sparsity in positive rewards and the curse of dimensionality, we propose a novel multi-agent HDRL framework by decomposing the decision process of the optimal portfolio weights into several sub-tasks. By deconstructing complex tasks into multiple sub-tasks, we can lower the difficulty of completing each sub-task, thereby facilitating the implementation of a divide-and-conquer strategy. Different agents are adopted in this novel framework to accomplish the sub-tasks. As a result, we reduce the training complexity of each sub-task in the training process to attain sufficient training of the neural networks in the DRL algorithm.

The key contributions of our work are as follows:

1) In our research, we introduce a novel tracking module to demonstrate that the strategies employed by our learning system in out-sample experiments stem from the thorough learning of market dynamics by its DRL agents during the training process. This module monitors indices that reflect the learning system's risk-adjusted profitability and risk control abilities throughout the training process. By exploring whether there are significant improvements in both the risk-adjusted profitability and risk control capabilities of our learning system during the training process, we can prove that the strategies executed by our learning system in out-sample experiments originate from the DRL agents' thorough exploration of optimal strategies to maximize the risk-adjusted return in the training environment.

2) We propose a novel learning system based on the HDRL algorithm for portfolio optimization. The in-sample performance proves that, by applying our multi-agent HDRL framework, we can adequately train the critic network to propagate gradient direction to the policy network that maximizes the reward function value. Consequently, the agents in the HDRL framework can effectively learn the policy for portfolio optimization to maximize the risk-adjusted return in the training environment while addressing the challenges imposed by the sparsity of positive reward and the curse of dimensionality.

3) We propose to adopt the Black-Litterman model based deep reinforcement learning agent (BDA) of Sun et al. [54] as the auxiliary agent to carry out the sub-task that assists in determining the optimal portfolio weights. In the back-testing experiments, the learning system demonstrates exceptional out-of-sample performance in profitability and return per unit risk by adopting such an auxiliary agent. In the back-test experiments, our portfolio management learning system outperforms the comparative strategies by a margin of at least 19.4% in terms of the Sharpe ratio [55] and Sortino ratio [56] while maintaining outstanding performance in profitability compared to these comparative strategies. Furthermore, empirical results from the ablation studies show that our portfolio management learning system can outperform each individual DRL agent in our multi-agent HDRL algorithm framework by a margin of 10% in terms of the Sharpe ratio and Sortino ratio. This suggests that the performance of the portfolio management learning system based on the HDRL algorithm can achieve state-of-the-art out-of-sample performance in maximizing the return per unit risk.

The remainder of this paper is organized as follows. Section 2 describes the preliminary work and defines the portfolio optimization problem addressed in this paper. Section 3 introduces the design of the HDRL framework in detail. In section 4, we compare the performance of our proposed HDRL algorithm to other contrasting strategies on multiple experimental back-testing datasets. Additionally, a detailed description of the results from the ablation experiments is also provided. Finally, Section 5 gives the conclusion and future work.

## 2. Preliminaries
### 2.1 Basic Financial Concepts

**Definition 2.1.1 (Trading period)** A trading period is the minimum time unit for reallocating funds among assets in a portfolio. Here, funds reallocation actions are executed at the end of the last trading day of each trading period. According to Li et al. [57], high-frequency stock trading data always exhibits a low signal-to-noise ratio. To prevent the DRL agent from learning too much noisy information from the environment, we extend the trading period to decrease the trading frequency. Hence, starting from the initial fund allocation action, the fund reallocation actions are executed every $K$ trading days throughout the investment horizon.

As given in Figure 1, the $t^{\text{th}}$ trading period is defined as the period interval $[t, t+1)$, $t = 1, 2, \ldots, T_f$, where $T_f$ is the total number of trading periods. Specifically, a trading period is a time period which begins when a fund's reallocation action is executed and ends just before the next fund reallocation action occurs. In a trading period, there are $K$ trading days. The $k^{\text{th}}$ trading day in the $t^{\text{th}}$ trading period is denoted as $t_k$, $k = 1, 2, \ldots, K$.

**Definition 2.1.2 (Closing price)** The adjusted closing price is used to measure the value of the stocks in the portfolio. The adjusted closing price refers to the finalized trading price of a stock that has undergone certain adjustments to account for various corporate behaviors such as dividends, stock splits, and other significant events [58]. The advantage of utilizing the adjusted closing price is its ability to provide a more accurate representation of the true performance of a stock over time. The closing price vector $\boldsymbol{p_t}$ is the adjusted closing price of the stocks in the portfolio at the last trading day within the $t^{\text{th}}$ trading period, i.e.,

$$\boldsymbol{p_t} = [p_{1,t}, p_{2,t}, \ldots, p_{n,t}]^T, \qquad (1)$$

where $p_{i,t}$ represents the adjusted closing price of stock $i$ on the last trading day in the $t^{\text{th}}$ trading period, and $n$ denotes the number of assets in the portfolio. To better describe the return distribution of the socks, we observe the adjusted closing price at the end of each trading day in the trading period. The adjusted closing price collected on the $k^{\text{th}}$ trading day in the $t^{\text{th}}$ trading period is defined as

$$\boldsymbol{p_{t_k}} = [p_{1,t_k}, p_{2,t_k}, \ldots, p_{n,t_k}]^T. \qquad (2)$$

**Definition 2.1.3 (Target portfolio weights)** The portfolio weights vector $\boldsymbol{w_t}$ represents the target portfolio weights at the beginning of the $t^{\text{th}}$ trading period, which are defined as

$$\boldsymbol{w_t} = [w_{1,t}, w_{2,t}, \ldots, w_{n,t}]^T,$$

where the $i^{\text{th}}$ component $w_{i,t}$ represents the ratio of the total portfolio value (money) invested in stock $i$ at the beginning of the $t^{\text{th}}$ trading period, and $n$ is the number of risk assets (i.e., stocks) in the portfolio. Furthermore, $w_{0,t} = 1 - \sum_{i=1}^{n} w_{i,t}$ is defined as the target weight of the risk-free asset (i.e., cash) at the beginning of the $t^{\text{th}}$ period. As defined in Definition 2.1.1, a trading period starts with asset reallocation carried out at the end of the trading days. Hence, the target portfolio weights $\boldsymbol{w_t}$ should be determined at the beginning of the $t^{\text{th}}$ trading period.

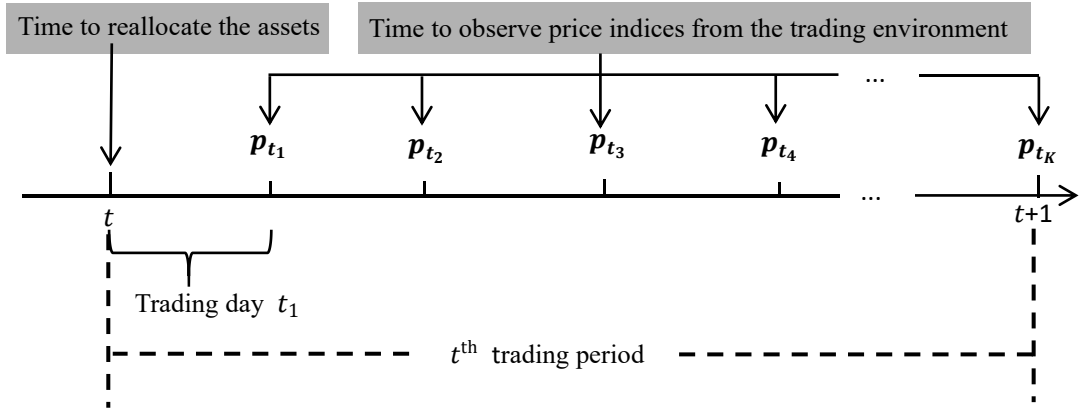

**Figure 1:** The designated observation intervals within the $t^\text{th}$ trading period. The $t^\text{th}$ trading period encompasses $K$ distinct trading days $t_k, k = 1, 2, \ldots K$. At the end of each trading day $t_k, k = 1, 2, \ldots, K$, the adjusted closing price vector $\boldsymbol{p}_{t_k}, k = 1, 2, \ldots, K$ is observed for analysis.

**Definition 2.1.4 (Target position)** The target position refers to the anticipated number of shares held during a given trading period. Based on the target portfolio weights $\boldsymbol{w}_t$ stated in Definition 2.1.2, we can calculate the target position $\boldsymbol{q}_t$ when determining the total investment $T_t$:

$$\boldsymbol{q}_t = [q_{1,t}, q_{2,t}, \ldots, q_{n,t}]^T = \lfloor (T_t \boldsymbol{w}_t) \oslash \boldsymbol{p}_{t-1} \rfloor, \tag{3}$$

where $\oslash$ represents the element-wise division. Under the assumption that one share of stock cannot be subdivided, we apply the floor function $\lfloor \cdot \rfloor$ in Equation (3) to make the target position $\boldsymbol{q}_t$ an integer.

To ensure the stability of the open position size in consecutive trading periods, the total investment scale is kept constant in each training and back-testing environment. Here, the total investment scale is set to the fixed ratio of the initial investment amount:

$$T_t = \eta T_0, \quad \text{for } t = 1, \ldots, T_f,$$

where $\eta \in (0, 1]$ is the investment ratio value. The value of the investment ratio $\eta$ is specified based on the investor's subjective view.

**Corollary 2.1.4.1** To hold the target position $\boldsymbol{q}_t$ of stocks, long position trade orders or short selling orders should be made to reallocate funds among portfolio assets. The market order $\Delta \boldsymbol{q}_t$ is defined to describe the modified stock positions at the beginning of the $t^\text{th}$ trading period, which can be calculated as

$$\Delta \boldsymbol{q}_t = [\Delta q_{1,t}, \Delta q_{2,t}, \ldots, \Delta q_{n,t}]^T = \boldsymbol{q}_t - \boldsymbol{q}_{t-1}. \tag{4}$$

**Definition 2.1.5 (Risk-free asset value)** Risk-free asset value $c_t$ is the value of the risk-free asset (i.e., cash) after the long/short operation at the beginning of the $t^\text{th}$ trading period, which is calculated as

$$c_t = \begin{cases} c_{t-1} - \Delta \boldsymbol{q}_t^T \boldsymbol{p}_{t-1} - \alpha |\Delta \boldsymbol{q}_t^T| \boldsymbol{p}_{t-1} - r_s \sum_{q_{t,i}<0} |q_{i,t}| p_{i,t-1} & c_{t-1} \geq 0 \\ c_{t-1}(1 + r_l) - \Delta \boldsymbol{q}_t^T \boldsymbol{p}_{t-1} - \alpha |\Delta \boldsymbol{q}_t^T| \boldsymbol{p}_{t-1} - r_s \sum_{q_{t,i}<0} |q_{i,t}| p_{i,t-1} & c_{t-1} < 0 \end{cases}, \tag{5}$$

where $\Delta \boldsymbol{q}_t$ is the market order defined in Equation (4), and $\boldsymbol{p}_{t-1}$ is the price vector defined in Equation (1). In Equation (5), $r_l$ is the borrowing rate of the risk-free asset, $r_s$ is the borrowing rate of the risk asset, and $\alpha$ is the commission fee rate of the transaction. Here, cash is the risk-free asset in the portfolio. When the value of the risk-free asset is larger than zero, it indicates that there is still cash left after making the target long/short operation. When the value of the risk-free asset is less than zero, it implies that the investor is in debt during this trading period.

**Definition 2.1.6 (Total assets value)** Total assets value is the total value of the risk-free asset and risk assets. The total assets value just before the asset reallocation at the end of the $t^\text{th}$ trading period is defined as $v_t$, which is calculated as:

$$v_t = c_t + \boldsymbol{q}_t^T \boldsymbol{p}_t, \tag{6}$$

where $c_t$ is the risk-free asset value defined in Equation (5). In Equation (6), $\boldsymbol{q}_t$ and $\boldsymbol{p}_t$ are as defined in Equation (3) and (1). In the $t^\text{th}$ trading period, the value of total assets at the end of the $k^\text{th}$ trading day is defined as $v_{t_k}$, which can be calculated as

$$v_{t_k} = c_t + \boldsymbol{q}_t^T \boldsymbol{p}_{t_k}, \tag{7}$$

where $\boldsymbol{p}_{t_k}$ is the closing price vector defined in Equation (2). In Equation (7), $c_t$ and $\boldsymbol{q}_t$ are as defined in Equation (3) and (1).

**Definition 2.1.7 (Logarithm return of the total assets)** To describe the changes in the value of total assets $v_t$, we define the logarithm return of the total assets at the end of the $t^{\text{th}}$ trading period as

$$\varpi_t = \log_2(v_t/v_{t-1}), \tag{8}$$

where $v_t$ is the value of total assets defined in Equation (6). In the $t^{\text{th}}$ trading period, the logarithm return of the total assets at the $k^{\text{th}}$ trading day is defined as $\varpi_{t_k}$, which can be calculated as

$$\varpi_{t_k} = \begin{cases} \log_2(v_{t_k}/v_{t_{k-1}}) & k = 2, 3, \ldots, K \\ \log_2(v_{t_k}/v_{t-1}) & k = 1 \end{cases},$$

where $v_{t_k}$ is the value of the total assets defined in Equation (7).

**Definition 2.1.8 (Portfolio value)** Due to the setup in our scenario, the investment amount $T_t$ is not equal to the total asset value $v_t$. Therefore, the portfolio value $v_t^p$ and total asset value $v_t$ are not equal. Since the risk-free asset value is fixed when there is no asset reallocation, the changes in the risk assets value in a trading period are equal to the changes in the total assets value. Based on the changes in the total assets value, the value of the portfolio in the trading period can be calculated. The value of the portfolio just before the asset reallocation at the end of the $t^{\text{th}}$ trading period is defined as $v_t^p$, which can be calculated as

$$v_t^p = v_t - v_{t-1} + T_t.$$

In the $t^{\text{th}}$ trading period, the value of the portfolio at the end of the $k^{\text{th}}$ trading day is defined as $v_{t_k}^p$, which can be calculated as

$$v_{t_k}^p = v_{t_k} - v_{t-1} + T_t.$$

**Definition 2.1.9 (Logarithm return of the portfolio)** When the target portfolio weights $w_t$ for asset reallocation is made at the end of the $t - 1^{\text{th}}$ trading, the logarithm return of the portfolio within the $t^{\text{th}}$ trading period $\xi_t$ can be calculated based on the adjusted closing price $p_t$ at the end of the $t^{\text{th}}$ trading period:

$$\xi_t = \log_2(v_t^p/T_t) = \log_2\left(\frac{c_t + \boldsymbol{q}_t^T \boldsymbol{p}_t - v_{t-1} + T_t}{T_t}\right). \tag{9}$$

For each trading day $k$ in the $t^{\text{th}}$ trading period, the daily logarithmic return of the portfolio $\vartheta_{t_k}$ can be calculated as

$$\vartheta_{t_k} = \begin{cases} \log_2(v_{t_k}^p/v_{t_{k-1}}^p) & k = 2, 3, \ldots, K \\ \log_2(v_{t_k}^p/v_{t-1}^p) & k = 1 \end{cases}.$$

**Definition 2.1.10 (Return vector)** Return vector comprises the log-return of the stocks in the portfolio. Since the return distribution is critical for the determination of the portfolio weights, we calculate the periodical return vector $\boldsymbol{R}_t$ at the end of the $t^{\text{th}}$ trading period and the daily return vector $\boldsymbol{z}_{t_k}$ on the $k^{\text{th}}$ trading day in the $t^{\text{th}}$ trading period. The periodical return vector $\boldsymbol{R}_t$ at the $t^{\text{th}}$ trading period is defined as

$$\boldsymbol{R}_t = [\log_2(p_{1,t}/p_{1,t-1}), \log_2(p_{2,t}/p_{2,t-1}), \ldots, \log_2(p_{n,t}/p_{n,t-1})]^T$$
$$\triangleq [R_{1,t}, R_{2,t}, \ldots, R_{n,t}]^T.$$

The daily return vector $\boldsymbol{z}_{t_k}$ on the $k^{\text{th}}$ trading day in the $t^{\text{th}}$ trading period is defined as

$$\boldsymbol{z}_{t_k} = [z_{1,t_k}, z_{2,t_k}, \ldots, z_{n,t_k}]^T$$
$$= \begin{cases} \log_2(\boldsymbol{p}_{t_k} \oslash \boldsymbol{p}_{t_{k-1}}) & k = 2, 3, 4, 5 \\ \log_2(\boldsymbol{p}_{t_k} \oslash \boldsymbol{p}_{t-1}) & k = 1 \end{cases}, \tag{10}$$

where the element $z_{i,t_k}$ is the log-return of stock $i$ at the $k^{\text{th}}$ trading day in the $t^{\text{th}}$ trading period. Here, the mean of the daily return vector $\boldsymbol{z}_{t_k}$ in the $t^{\text{th}}$ trading period is defined as

$$\boldsymbol{\mu}_t = [\mu_{1,t}, \mu_{2,t}, \ldots, \mu_{n,t}]^T,$$

where $\mu_{i,t}$ is the expectation of the daily return of stock $i$ on the $t^{\text{th}}$ trading period, which is calculated by

$$\mu_{i,t} = \frac{1}{K} \sum_{k=1}^{K} z_{i,t_k}.$$

The covariance matrix of the daily return vector $\boldsymbol{z}_{t_k}$ in the $t^{\text{th}}$ trading period is defined as

$$\Sigma_t = \begin{bmatrix} \sigma_{11,t}^2 & \cdots & \sigma_{1n,t}^2 \\ \vdots & \ddots & \vdots \\ \sigma_{n1,t}^2 & \cdots & \sigma_{nn,t}^2 \end{bmatrix}, \tag{11}$$

where $\sigma_{ii,t}^2$ is the variance of stock $i$ in the $t^{\text{th}}$ trading period and $\sigma_{ij,t}^2$ is the covariance between stock $i$ and stock $j$ in the $t^{\text{th}}$ trading period. In our research, the calculation of the variance $\sigma_{ii,t}^2$ and covariance $\sigma_{ij,t}^2$ is based on the return data in the $t^{\text{th}}$ trading period and prior $M-1$ trading periods, which is calculated as:

$$\sigma_{ii,t}^2 = \frac{\sum_{m=0}^{M-1}\sum_{k=1}^{K}\left[z_{i,\,t-m_k} - \frac{1}{KM}\sum_{m=0}^{M-1}\sum_{k=1}^{K}z_{i,\,t-m_k}\right]^2}{KM-n-1}; i = 1, 2, \ldots, n$$

$$\sigma_{ij,t}^2 = \frac{\sum_{m=0}^{M-1}\sum_{k=1}^{K}\left[z_{i,\,t-m_k} - \frac{1}{KM}\sum_{m=0}^{M-1}\sum_{k=1}^{K}z_{i,\,t-m_k}\right]\left[z_{j,\,t-m_k} - \frac{1}{KM}\sum_{m=0}^{M-1}\sum_{k=1}^{K}z_{j,\,t-m_k}\right]}{KM-n-1}; i \neq j$$

**Corollary 2.1.10.1 (Price fluctuation tensor)** To provide a more comprehensive description of the state for the DRL agents, we construct a price fluctuation tensor $Y_t \in \mathbb{R}^{K \times n}$ that represents the historical distribution of returns for the portfolio assets. Price fluctuation tensor $Y_t$ is a matrix which comprises all the daily return vectors in the $t^{\text{th}}$ trading period. The price fluctuation tensor at the $t^{\text{th}}$ trading period is denoted as

$$Y_t = [z_{t_1}, z_{t_2}, \ldots, z_{t_K}], \tag{12}$$

where $z_{t_k} \in \mathbb{R}^{n \times 1}, (k = 1, 2, \ldots, K)$ is the daily return vector defined in Equation (10).

**Definition 2.1.11 (Transaction scale)** In portfolio optimization in consecutive trading periods, the transaction cost is a significant factor in the calculation of the portfolio return. Hence, we try to control the transaction scale $o_t$, which is calculated as:

$$o_t = |\Delta q_t|^T p_{t-1},$$

where $\Delta q_t$ is the market order defined in Equation (4), $p_{t-1}$ is the price vector defined in Equation (1). In market order $\Delta q_t$, the number of stocks bought and sold are respectively expressed by positive and negative numbers. Hence, we use the absolute values of market order, i.e., $|\Delta q_t|$, to calculate the transaction scale for both buy and sell orders. Furthermore, we define the transaction ratio $\epsilon_t$, which is the ratio between the scale of trading positions and the total open position scale:

$$\epsilon_t = \frac{|\Delta q_t|^T p_{t-1}}{T_t}, \tag{13}$$

where $T_t$ is the total investment amount at the $t^{\text{th}}$ trading period.

**Definition 2.1.12 (Variance)** Variance is a popular index for measuring the risk of the portfolio. Here, the variance of the portfolio $w_t$ is defined as $V_t$, which is calculated as

$$V_t = w_t^T \Sigma_t w_t, \tag{14}$$

where $\Sigma_t$ is the covariance matrix defined in Equation (11).

## 2.2 Trading process

As given in Figure 2, the target portfolio weights $w_t$ are determined at the beginning of the $t^{\text{th}}$ trading period, and the positions of each portfolio asset are adjusted at the beginning of the $t^{\text{th}}$ trading period by the following steps:

(1) Calculate the target position $q_t$ based on Equation (3). If the target position for stock $i$: $q_{i,t} \geq 0$, it means that investors should hold $q_{i,t}$ shares of the stock $i$ in the long position at the $t^{\text{th}}$ trading period. If $q_{i,t} < 0$, the investor should hold $-q_{i,t}$ shares of stock $i$ in the short sale.

(2) Calculate the market order vector $\Delta q_t$ based on Equation (4). For stock $i, \{i = 1, 2, \ldots, n\}$ in the portfolio, investors should make the long position trade orders or short selling orders based on the stocks held in hand. The specific operations carried out by the agents are summarized in Table 1.

| Value of the market order | Relationship of the market order and stock positions held in hand | Investment operations to be carried out at the beginning of the $t^{\text{th}}$ trading period |
| --- | --- | --- |
| $\Delta q_{i,t} \geq 0$ | $q_{i,t-1} \geq 0$ | Buy $\Delta q_{i,t}$ shares of stocks. |
| | $q_{i,t-1} \leq 0$, and $|q_{i,t-1}| \geq \Delta q_{i,t}$ | Buy $\Delta q_{i,t}$ shares of stocks and return such stocks. |
| | $q_{i,t-1} \leq 0$, and $|q_{i,t-1}| < \Delta q_{i,t}$ | Buy $\Delta q_{i,t}$ shares of stock and return $|q_{i,t-1}|$ shares of stocks. |
| $\Delta q_{i,t} < 0$ | $q_{i,t-1} \leq 0$ | Borrow $|\Delta q_{i,t}|$ shares of stock and sell them. |
| | $q_{i,t-1} > 0$ and $q_{i,t-1} \geq |\Delta q_{i,t}|$ | Sell $|\Delta q_{i,t}|$ shares of stocks held in hand. |
| | $q_{i,t-1} > 0$ and $q_{i,t-1} < |\Delta q_{i,t}|$ | Sell $q_{i,t-1}$ shares of stocks hold in hand and borrow $|\Delta q_{i,t}| - q_{i,t-1}$ shares of stocks to sell it. |

**Table 1.** The specific operation summary

*2.3 Problem setting*

In this research, the objective is to maximize the return per unit risk by determining the target portfolio weights $\boldsymbol{w_t}, t = 1, 2, \ldots, T_f$ and adjusting the position of each portfolio asset at the beginning of the $t^{\text{th}}$ ($t = 1, 2, \ldots, T_f$) trading periods.

*2.4 Assumptions*

This study relies exclusively on back-testing to conduct all experiments, whereby the investors commence trading at a specific historical time-point with no prior knowledge of future market conditions. To satisfy the requirements of the back-test tradings set in each experiment, we make the following assumptions:

(A1)  Zero Slippage: The trading decisions executed at the beginning of each trading period are assumed to have no impact on the asset price movement.

(A2) Zero Market Impact: The assets included in the portfolio are characterized by high levels of market liquidity, thereby enabling immediate execution of all the trading orders upon the orders being issued.

(A3)  Short Selling Permit: There are no limitations imposed on short-selling activities in the target market.

In a practical trading setting, the aforementioned assumptions (A1) and (A2) hold true, provided that the trading volume of the selected assets in the portfolio is substantial. To satisfy assumption (A3), we conduct experiments within the stock market, allowing for long trades and short-selling trades (i.e., the United States stock market).

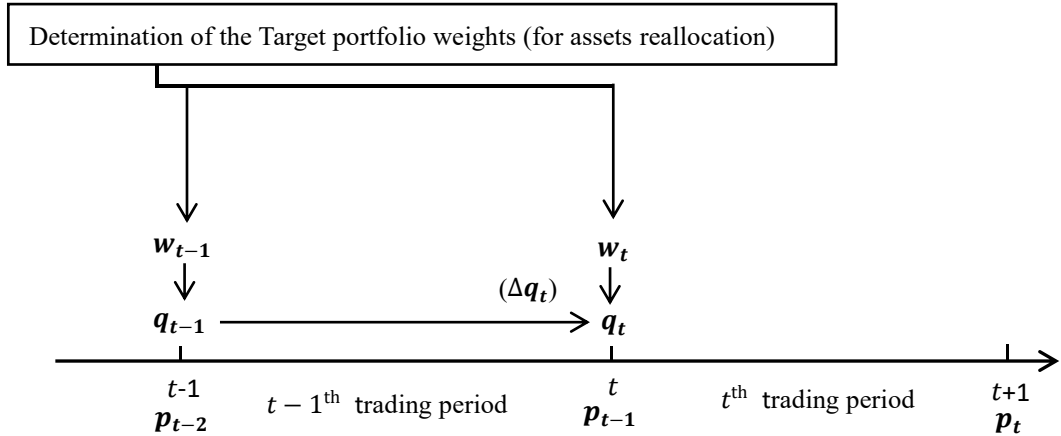

**Figure 2:** The illustration of the trading process in each trading period. The assignment of target weights $\boldsymbol{w_t}$ to a portfolio for the purpose of asset reallocation should be determined at the beginning of the $t^{\text{th}}$ trading period. Based on the target weights $\boldsymbol{w_t}$, the corresponding target position $\boldsymbol{q_t}$ and market order $\Delta \boldsymbol{q_t}$ can be derived. Subsequently, the market order $\Delta \boldsymbol{q_t}$ is executed to realize the target portfolio weights $\boldsymbol{w_t}$.

# 3. Methodology

As part of the HDRL framework, different DRL agents engage in iterative interactions with an uncertain environment and update their policies to converge to optimal policies. The dynamic decision-making process of different agents fits naturally into the framework of the finite Markov Decision Process (MDP), which is defined by the tuple:

$$MDP = <\mathcal{S}, \mathcal{A}, \mathcal{P}, \mathcal{R}, \gamma, \mathcal{T}>,$$

where $\mathcal{S}$ represents the state space and $\mathcal{A}$ represents the action space. The state and action space are continuous in the Markov Decision process. $\mathcal{P}: \mathcal{S} \times \mathcal{A} \rightarrow \mathcal{S}$ is the state transition dynamics, defined as a probability distribution $P(s_{t+1}|s_t, a_t)$. It represents the probability of the successor state $s_{t+1}$ when given the state $s_t \in \mathcal{S}$ and taking the action $a_t \in \mathcal{A}$. $\mathcal{R}: \mathcal{S} \times \mathcal{A} \rightarrow \mathbb{R}$ is the reward function. $\gamma$ is the discount factor, and $\mathcal{T}$ is the set of terminal states.

### 3.1 The training objective of the learning system.

Our learning system's training objective is to maximize accumulated return and control the risk simultaneously. According to this objective, we construct the objective function $\Theta(w_t)$ to evaluate the target portfolio weights $w_t$ determined at the beginning of the trading period $t$. To achieve a balance between the expected return and risk of the portfolio, the objective function for the DRL agents in portfolio optimization is formulated as the return adjusted for risk and the corresponding transaction scale. Specifically, the objective function $\Theta(w_t)$ is defined as

$$\Theta(w_t) = \xi_t/K - \lambda_1 V_t, -\lambda_2 \epsilon_t, \qquad (15)$$

where $\xi_t$ is the logarithm return of the investment based on the portfolio weights $w_t$ (Equation (9)), and the term $\xi_t/K$ measures the daily logarithm return of the investment. In Equation (15), $V_t$ and $\epsilon_t$ are, respectively, the variance of the target portfolio weights $w_t$ and the transaction scale ratio of the corresponding market order. To control the risk of the portfolio and avoid high transaction fees, the variance $V_t$ and the transaction scale ratio $\epsilon_t$ of the target portfolio $w_t$ should be controlled. Hence, the values of the parameters $\lambda_1$ and $\lambda_2$ should be larger than zero.

The training objective of the learning system is to learn a policy function $\pi_\phi$, which determines the target portfolio weights $w_t$ at each trading period with an objective that the expectation of objective function $\Theta(w_t)$ at each trading period can be maximized, i.e.,

$$\underset{\pi_\phi}{\arg\max} \, \mathbb{E}_{w_t \sim \pi_\phi, t=1, 2, \ldots, T_f}[\Theta(w_t)].$$

### 3.2 The Design of Two Agents

The proposed hierarchical DRL algorithm framework consists of both the auxiliary agent and the executive agent. The sub-task of the auxiliary agent is to explore the baseline portfolio weights $w_t^{au}$, which can obtain positive objective function $\Theta(w_t)$ value from the environment. By providing the baseline portfolio weights $w_t^{au}$ for the executive agent, the auxiliary agent assists the executive agent in completing the asset reallocation. Hence, the sub-task of the auxiliary agent is an auxiliary task [59]. The sub-task of the executive agent is to explore the optimal policy which can maximize the objective function $\Theta(w_t)$ and ensure the generalization capability of policy simultaneously. Based on the baseline portfolio weights $w_t^{au}$ to explore the optimal policy, the executive agent can avoid exploring most of the action space with negative logarithm return and large variance. In this way, it is anticipated that the accumulative value of risk-adjusted returns obtained by our learning system from the training environment can be significantly improved during the training process.

### 3.3 Hierarchical MDP Framework

Based on the method of Wang et al. [60], the decision-making process of the auxiliary agent and executive agent in the environment is modeled as two hierarchical MDPs [60]. In this section, we define the components in these two MDPs for the auxiliary agent and executive agents, respectively. The decision process of the auxiliary agent is modeled as the auxiliary-level MDP:

$$MDP^{au} = <\mathcal{S}^{au}, \mathcal{A}^{au}, \mathcal{P}^{au}, \mathcal{r}^{au}, \gamma^{au}, \mathcal{T}^{au}>,$$

and the decision process of the executive agent is modeled as the executive-level MDP:

$$MDP^{ex} = <\mathcal{S}^{ex}, \mathcal{A}^{ex}, \mathcal{P}^{ex}, \mathcal{r}^{ex}, \gamma^{ex}, \mathcal{T}^{ex}>,$$

where the superscripts a u and e x in $MDP^{au}$ and $MDP^{ex}$ represent the auxiliary agent and the executive agent, respectively. Since the two agents work together to determine the target portfolio at the beginning of each trading period, the set of terminal states $\mathcal{T}^{au}$ and $\mathcal{T}^{ex}$ in the auxiliary-level MDP and executive-level MDP are the same, which is the termination conditions of the asset reallocation in the consecutive trading period. Here, the termination condition is defined as the current trading period's serial number $t$ larger than the number of the

$$t > T_f.$$

The detailed definitions of the remaining components in two hierarchical MDPs are given in the following sub-section.

**The auxiliary-level MDP for the auxiliary agent**

In the learning system, the sub-task of the auxiliary agent involves determining the baseline portfolio weights $w_t^{au}$ and assisting the executive agent in determining the portfolio weights for asset reallocation. In this way, the executive agent can avoid exploring the optimal policy in the action space with negative logarithm returns and larger variance. Hence, it is expected that the baseline portfolio weights determined by the auxiliary agent can obtain the positive objective function value from the environment. Moreover, since the executive agent determines the portfolio weights $w_t^{ex}$ for asset reallocation based on the baseline portfolio weights $w_t^{au}$, the generalization capability of the auxiliary agent should be ensured. Here, we adopt the BL-based DRL agent (BDA) proposed by Sun et al. [54] to perform the task of the auxiliary agent because it is able to effectively utilize the long/short strategy that maximizes the return per unit risk. Moreover, it has demonstrated outstanding out-of-sample performance in terms of returns per unit risk in the U.S. stock market. The components in the auxiliary-level MDP are based on Sun et al.'s [54] idea. The detailed definitions of the components in the auxiliary-level MDP are given as follows:

**State $s_t^{au}$.** The state $s_t^{au} \in S^{au}$ in the auxiliary-level MDP is described with a tuple consisting of the historical return tensor $X_t$ and portfolio weights $w_{t-1}^{au}$ determined by the auxiliary agent at the prior trading period:

$$s_t^{au} = \langle w_{t-1}^{au}, X_t \rangle.$$

The historical return tensor $X_t$ is constructed using the price fluctuation tensor $Y_t$ defined in Equation (12) by a time window $M$:

$$X_t = [Y_{t-M}, Y_{t-M+1}, \ldots, Y_{t-1}].$$

Since the description of the state $s_t^{au}$ needs the price fluctuation tensor $Y_t$ in the past $M$ trading period, we should observe the adjusted closing price $p_t$ in the past $KM + 1$ trading days to satisfy the requirement for data. Hence, in the training environment, the first trading period is set at the end of the $KM + 1$ trading days. In the back-test experiments, we collect the price fluctuation tensor $Y_t$ for $M$ trading periods in advance before the back-test trading begins.

**Action $a_t^{au}$.** Based on the environment state $s_t^{au}$, the auxiliary agent outputs the baseline portfolio weights $w_t^{au}$, which subsequently becomes input into the executive agent to assist in asset reallocation. Hence, the action $a_t^{au} \in A^{au}$ of the auxiliary agent is defined as:

$$a_t^{au} = w_t^{au} = [w_{1,t}^{au}, w_{2,t}^{au}, \ldots, w_{n,t}^{au}]^T,$$

where $w_{i,t}^{au}$ is the weight of asset $i$ in the portfolio at the $t^{th}$ trading period, and $n$ is the number of portfolio assets. The superscript $au$ in the symbols $a_t^{au}$ and $w_t^{au}$ represents the auxiliary agent.

**Reward $r^{au}$.** Since the policy function in the BDA is updated based on the policy gradient algorithm of Jiang et al. [20], the reward function should be defined as a continuously differentiable function of the portfolio weights. The reward function $r^{au}$ is formulated using the following steps:

**Step 1)** A continuously differentiable evaluation function $\rho(a_t^{au}, s_t^{au} | \lambda_1, \lambda_2)$ is constructed based on the objective function $\Theta(w_t)$ defined in Equation (15):

$$\rho(a_t^{au}, s_t^{au} | \lambda_1, \lambda_2) = w_t^{au T}\mu_t - \lambda_1 w_t^{au T}\Sigma_t w_t^{au} - \lambda_2 |w_t^{au} - w_{t-1}^{au}|^T e,$$

where the term $w_t^{au T}\mu_t$ is the estimation of daily logarithm return expectation, which is calculated by the term $\xi_t/K$ in Equation (15); the term $w_t^{au T}\Sigma_t w_t^{au}$ reflects the variance $V_t$ defined in Equation (14). In the term $|w_t^{au} - w_{t-1}^{au}|^T e$, the vector $e \in \mathbb{R}^{N \times 1}$ is a vector of all ones. Hence, $|w_t^{au} - w_{t-1}^{au}|^T e$ is the estimate of transaction scale ratio $\epsilon_t$ defined in Equation (13). The parameters $\lambda_1$ and $\lambda_2$ are as defined in the objective function $\Theta(w_t)$.

**Step 2)** A target value is set for the evaluation function $\rho(a_t^{au}, s_t^{au} | \lambda_1, \lambda_2)$ to avoid overfitting, leading to the generalization ability of the policy. The target value is calculated based on Markowitz's theory [2] (1959), which is the evaluation function $\rho(a_t^{au}, s_t^{au} | \lambda_1, \lambda_2)$ value of the theoretically optimal portfolio weights $w_t^{optimal}$ under the unconstrained portfolio optimization problem with the risk aversion $\lambda_3$:

$$\begin{cases} \Gamma_t = \rho(w_t^{optimal}, s_t^{au} | \lambda_1, \lambda_2) = w_t^{optimal T}\mu_t - \lambda_1 w_t^{optimal T}\Sigma_t w_t^{optimal} - \lambda_2 |w_t^{optimal} - w_{t-1}^{au}|^T e \\ w_t^{optimal} = \left(\frac{1}{\lambda_3}\right)\Sigma_t^{-1}\mu_t \end{cases}.$$

**Step 3)** Based on the evaluation function $\rho(w_t, s_t^{au} | \lambda_1, \lambda_2)$ and the target value $\Gamma_t$, the reward $r^{au}$ for the state-action pair $(s_t^{au}, a_t^{au})$ in the $t^{th}$ trading period is defined as :

$$r^{au}: \begin{cases} r_t^{au} = r^{au}(a_t^{au}, s_t^{au}| \lambda_1, \lambda_2, \lambda_3) = -[\Gamma_t - \rho(a_t^{au}, s_t^{au}| \lambda_1, \lambda_2)]^2 \\ \Gamma_t = \rho(w^{optimal}, s_t^{au}| \lambda_1, \lambda_2) \\ w^{optimal} = \left(\frac{1}{2\lambda_3}\right) \Sigma_t^{-1} \mu_t \end{cases} \quad (16)$$

The design of the reward function $r^{au}$ defined in Equation (16) allows the evaluation function $\rho(w_t, s_t^{au}| \lambda_1, \lambda_2)$ to converge to this target value $\Gamma_t$.

**The executive-level MDP for the executive agent**

In the learning system, the sub-task of the executive agent is to determine the target portfolio weights $w_t$ for asset reallocation with the assistance of the auxiliary agent. Based on the task of the executive agent and the training objective of the learning system, the components in the executive-level MDP are defined as follows:

**State $s_t^{ex}$.** The state $s_t^{ex} \in \mathcal{S}^{ex}$ in the executive-level MDP is described with the historical return tensor $X_t$ and the baseline portfolio weights $w_t^{au}$ determined by the auxiliary agent:
$$s_t^{ex} = \langle w_t^{au}, X_t \rangle.$$

**Action $a_t^{ex}$.** Since the task of the executive agent is to determine the portfolio weights $w_t^{ex}$ for asset reallocation, the action of the executive agent $a_t^{ex} \in \mathcal{A}^{ex}$ is defined as:
$$a_t^{ex} = w_t^{ex} = [w_{1,t}^{ex}, w_{2,t}^{ex}, \ldots, w_{n,t}^{ex}]^T,$$
where $w_{i,t}^{ex}$ is the weight of asset $i$ determined by the executive agent at the $t^{th}$ trading period. The portfolio weights $w_t^{ex}$ determined by the executive agent are the target portfolio weights $w_t$ defined in Definition 2.1.3:
$$w_t^{ex} = w_t.$$

**Reward $r^{ex}$.** The objective function $\Theta(w_t)$ defined in Equation (15) is directly used to evaluate the action of the executive agent. Since there are both long and short positions in the portfolio $w_t^{ex}$, in the process of exploring the optimal policy, certain executive portfolio weights $w_t^{ex}$ may result in the total value of the portfolio $v_t^p$ falling below zero at the end of the $t^{th}$ trading period. It renders the calculation of logarithmic returns for the portfolio infeasible. To mitigate this, we set a lower bound B for the total portfolio value function $v_t^p$ in the reward $r^{ex}$ calculation. In the hierarchical DRL framework, the reward for the state-action pair $(s_t^{ex}, a_t^{ex})$ in the executive-level MDP is defined as a piece-wise function:

$$r^{ex}: r^{ex}(a_t^{ex}, s_t^{ex}| \lambda_1, \lambda_2) = \begin{cases} \frac{1}{K}\log_2(v_t^p/T_t) - \lambda_1 w_t^T \Sigma_t w_t - \lambda_2 \epsilon_t & v_t^p \geq B \\ \frac{1}{K}\log_2(B/T_t) - \lambda_1 w_t^T \Sigma_t w_t - \lambda_2 \epsilon_t & v_t^p < B \end{cases}. \quad (17)$$

*3.4 Hierarchy of Two Policies*

In the learning system, the policy function of the auxiliary agent and the executive agent is defined as the auxiliary policy $\pi_\phi^{au}(s_t^{au})$ and the executive policy $\pi_\theta^{ex}(s_t^{ex})$. These two policy functions are executed successively at the beginning of each trading period. According to the definition of the components in the two dynamics, we shall summarize the decision logic of the executive agent and the auxiliary agent at the beginning of each trading period. As given in Figure 3, the decision logic of the agents in the learning system determines the target portfolio weights $w_t$ and can be summarized as the following four steps:
- Step 1: The auxiliary agent observes the state $s_t^{au}$ and determines the baseline portfolio weights $w_t^{au}$ based on the policy $\pi_\phi^{au}$.
- Step 2: The baseline portfolio weights $w_t^{au}$ are adopted as the significant tutorial information for the executive agent in the exploration of the optimal policy.
- Step 3: The executive agent observes the state $s_t^{ex}$, which includes the baseline portfolio weights $w_t^{au}$ and historical return information $X_t$. Subsequently, it decides the executive portfolio weights $w_t^{ex}$ by modifying the baseline portfolio weights $w_t^{au}$ based on the policy $\pi_\theta^{ex}$.
- Step 4: The executive portfolio weights are the target portfolio weights $w_t$, and the executive agent makes asset reallocation based on the executive portfolio weights $w_t^{ex}$.

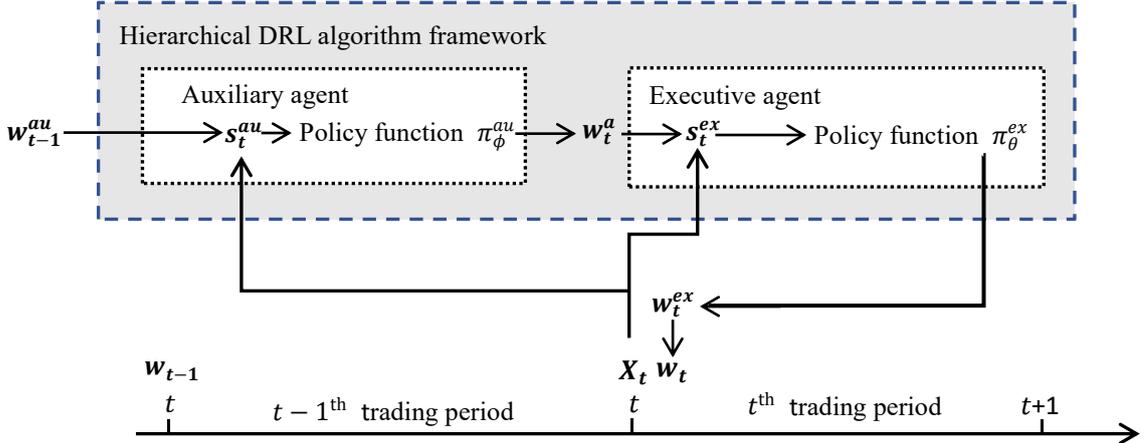

**Figure 3:** The decision process of DRL agents in the HDRL framework at the end of the $t-1^{th}$ trading period. The auxiliary agent first determines the baseline portfolio weights $w_t^{au}$ based on the historical return tensor $X_t$ and its action $w_{t-1}^{au}$ from the previous trading period. It then determines the baseline portfolio weights $w_t^{au}$ based on its policy function $\pi_\phi^{au}$ and inputs the baseline portfolio weights $w_t^{au}$ into the executive agent to assist it in determining the portfolio weights $w_t^{ex}$. Based on the baseline portfolio weights $w_t^{au}$ of the auxiliary agent and the historical return tensor $X_t$, the executive agent determines the portfolio weights $w_t^{ex}$ by its policy function $\pi_\theta^{ex}$. The portfolio weights $w_t^{ex}$ is the target portfolio weights $w_t$ for asset allocation.

*3.5 Optimizing via Hierarchical Deep Reinforcement Learning*

In this subsection, we provide a detailed description of the algorithm for training the DRL agents in the HDRL algorithm framework to learn hierarchical policies for portfolio optimization. The training logic framework of the DRL agents in the learning system is given in Figure 4. In the HDRL algorithm framework, the training objective of the learning system defined in Section 3.1 is broken down into the training objective of each agent. For the hierarchical policies of auxiliary agent and executive agent, the training objective of these two agents are respectively defined as:

$$\pi_\phi^{au*} = \underset{\pi_\phi}{\arg\max}\, \mathbb{E}_{s_1^{au} \sim \beta_1^{au},\, s_{t+1}^{au} \sim p^{au}(\cdot|s_t^{au}, a_t^{au}),\, a_t^{au} \sim \pi_\phi^{au}(s_t^{au})}[r_t^{au}]$$

$$\pi_\theta^{ex*} = \underset{\pi_\theta}{\arg\max}\, \mathbb{E}_{s_1^{ex} \sim \beta_1^{ex},\, s_{t+1}^{ex} \sim p^{ex}(\cdot|s_t^{ex}, a_t^{ex}),\, a_t^{ex} \sim \pi_\theta^{ex}(s_t^{ex})}[r_t^{ex}],$$

where $\beta_1^{au}$ and $\beta_1^{ex}$ denotes the initial auxiliary-level state distribution and initial executive-level distribution, respectively.

As given in Figure 4, the agents in the learning system are trained in two steps. First, the auxiliary agent explores the optimal policy independently based on the policy gradient algorithm. Second, after completing training the auxiliary agent, the executive agent explores the optimal policy for portfolio optimization based on the baseline portfolio weights.

Since directly letting the DRL agent explore the optimal strategies from the environment based on the actor-critic algorithm and the deep function approximators may lead to sparsity in positive reward and curse of dimensionality, the auxiliary agent in the learning system is trained by the off-policy deep policy gradient algorithm, based on Sun et al. [54]. In the training process, the objective $\mathcal{L}_{\pi^{au}}(\phi)$ of the auxiliary agent for learning the optimal policy function $\pi_\phi^{au*}(s_t^{au})$ is defined based on the continuously differentiable reward function defined in Equation (16):

$$\mathcal{L}_{\pi^{au}}(\phi) = \mathbb{E}_{(s_i^{au}, a_i^{au}, r_i^{au}, s_{i+1}^{au}) \sim B^{au}}[r^{au}(\pi_\phi^{au}(s_i^{au}), s_i^{au}|\lambda_1, \lambda_2, \lambda_3)],$$

where $B^{au}$ is the replay buffer with transitions of the executive agent, and $(s_i^{au}, a_i^{au}, r_i^{au}, s_{i+1}^{au})$ is the sample random extracted from the reply buffer $B^{au}$.

Since the reward function $r^{ex}(a_t^{ex}, s_t^{ex}|\lambda_1, \lambda_2)$ defined in Equation (17) is not a continuously differentiable function of the portfolio weights $w_t^{ex}$, we cannot directly use the reward function $r_t^{ex}$ to construct the objective function for the policy network. To achieve the training objective of maximizing the reward function $r_t^{ex}$, we use the actor-critic algorithm: deep deterministic policy gradient (DDPG) [42] to train the executive agent. In the decision logic of the executive agent, the executive portfolio weights $\boldsymbol{w}_t^{ex}$ are determined by modifying the baseline portfolio weights $\boldsymbol{w}_t^{au}$. Since baseline portfolio weights determined by the auxiliary agent can almost lead to positive rewards from the environment, executive portfolio weights $\boldsymbol{w}_t^{ex}$, which are modified from the baseline portfolio weights $\boldsymbol{w}_t^{au}$, are more likely to lend to positive returns. Hence, the DRL agent can focus on exploring the optimal policy within the action space with positive rewards from the environment. Therefore, the efficiency of optimal policy exploration can be enhanced when the actor-critic algorithm and deep function approximator are applied to train the DRL agent to carry out the task. Hence, we can avoid the issues of sparsity in positive reward and the curse of dimensionality in the training process. In the reinforcement learning algorithm, the action-value function $Q^{\pi_\theta^{ex}}(\boldsymbol{s}_t^{ex}, \boldsymbol{a}_t^{ex})$ is used to describe the expected return after executing action $\boldsymbol{a}_t^{ex}$ in the state $\boldsymbol{s}_t^{ex}$ and thereafter following the policy $\pi_\theta^{ex}$:

$$Q^{\pi_\theta^{ex}}(\boldsymbol{s}_t^{ex}, \boldsymbol{a}_t^{ex}) = \mathbb{E}_{\boldsymbol{s}_{i>t}^{ex}, r_{i\geq t}^{ex}, \boldsymbol{a}_{i>t}^{ex} \sim \pi_\theta^{ex}} \left[ \sum_{i=t}^T \gamma^{i-t} r^{ex}(\boldsymbol{s}_i^{ex}, \boldsymbol{a}_i^{ex}) \mid \boldsymbol{s}_t^{ex}, \boldsymbol{a}_t^{ex} \right].$$

Based on the action-value function $Q^{\pi_\theta^{ex}}(\boldsymbol{s}_t^{ex}, \boldsymbol{a}_t^{ex})$, the training objective of the executive agent is converted to learning the optimal policy:

$$\pi_\theta^{ex*} = \underset{\pi_\phi^{ex}}{\operatorname{argmax}} \, \mathbb{E}_{\boldsymbol{s}_{t+1}^{ex} \sim p^{ex}(\cdot|\boldsymbol{s}_t^{ex}, \boldsymbol{a}_t^{ex}), \boldsymbol{a}_t^{ex} \sim \pi_\phi^{ex}(\boldsymbol{s}_t^{ex})} \left[ \sum_{i=t}^T \gamma^{i-t} r^e(\boldsymbol{s}_i^{ex}, \boldsymbol{a}_i^{ex}) \mid \boldsymbol{s}_t^{ex}, \boldsymbol{a}_t^{ex} \right],$$

for $t = 1, 2, \ldots, T_f$.

In the DDPG algorithm, the critic network $Q_\omega(\boldsymbol{s}_t^{ex}, \boldsymbol{a}_t^{ex})$ is adopted to approximate the action-value function $Q^{\pi_\theta^e}(\boldsymbol{s}_t^{ex}, \boldsymbol{a}_t^{ex})$, i.e.,

$$Q_\omega(\boldsymbol{s}_t^{ex}, \boldsymbol{a}_t^{ex}) \approx \operatorname{Max}_{\pi_\theta^{ex}} Q^{\pi_\theta^{ex}}(\boldsymbol{s}_t^{ex}, \boldsymbol{a}_t^{ex}),$$

and the critic network $Q_\omega(\boldsymbol{s}_t^{ex}, \boldsymbol{a}_t^{ex})$ is learned by the one-step Bellman residual [85]. In the training process, the critic network $Q_\omega(\boldsymbol{s}_t^{ex}, \boldsymbol{a}_t^{ex})$ is optimized by minimizing a mean squared error loss:

$$\mathcal{L}_Q(\omega) = \mathbb{E}_{(\boldsymbol{s}_i^{ex}, \boldsymbol{a}_i^{ex}, r_i^{ex}, \boldsymbol{s}_{i+1}^{ex}) \sim B^{ex}} [Q_\omega(\boldsymbol{s}_i^{ex}, \boldsymbol{a}_i^{ex}) - (r_i^{ex} + \gamma \, Q_\omega(\boldsymbol{s}_{i+1}^{ex}, \pi_\theta^{ex}(\boldsymbol{s}_{i+1}^{ex})))]^2, \tag{18}$$

where $\gamma$ is the discount factor for the DRL agent to balance the short-term and long-term rewards. $B^{ex}$ is the replay buffer with transitions of the executive agent, and $(\boldsymbol{s}_i^{ex}, \boldsymbol{a}_i^{ex}, r_i^{ex}, \boldsymbol{s}_{i+1}^{ex})$ is the sample extracted from the replay buffer $B^{ex}$.

The policy function $\pi_\theta^{ex}(\boldsymbol{s}_t^{ex})$ is the actor in the DDPG algorithm, and the objective function used to update the policy function $\pi_\theta^{ex}(\boldsymbol{s}_t^{ex})$ of the executive agent is

$$\mathcal{L}_{\pi^{ex}}(\theta) = \mathbb{E}_{(\boldsymbol{s}_i^{ex}, \boldsymbol{a}_i^{ex}, r_i^{ex}, \boldsymbol{s}_{i+1}^{ex}) \sim B^{ex}} [Q_\omega(\boldsymbol{s}_i^{ex}, \pi_\theta^{ex}(\boldsymbol{s}_i^{ex}))]. \tag{19}$$

A detailed description of the training process is given in Algorithm 1, and the hyper-parameters in the HDRL algorithm are described in detail in Appendix C.

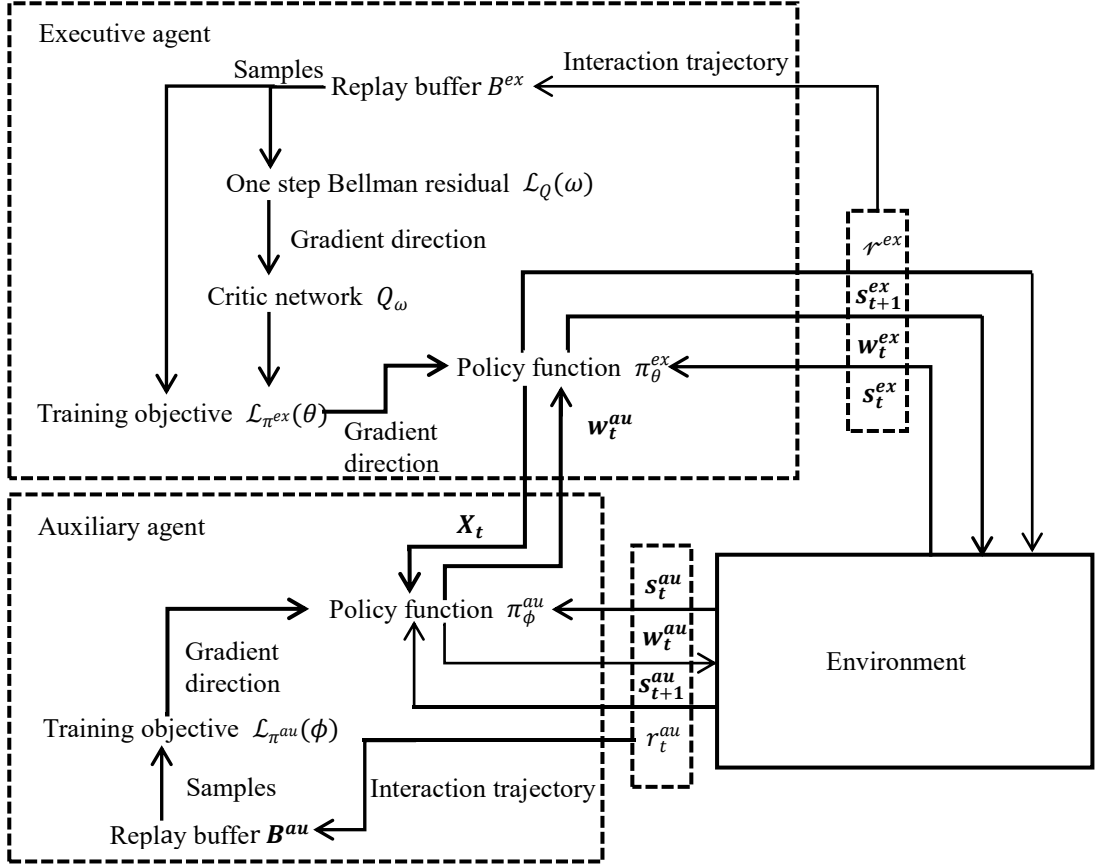

**Figure 4:** The training process of the agents in the HDRL algorithm framework can be divided into two phases. In the first phase, the auxiliary agent independently interacts with the financial environment and trains the policy function $\pi_\phi^{au}$ based on its policy function's training objective $\mathcal{L}_{\pi^{au}}(\phi)$ and interaction trajectory samples in the replay buffer $B^{au}$. In the second phase, the auxiliary agent and the trained executive agent work together to interact with the financial environment. The interaction trajectory is stored in the replay buffer $B^{ex}$. Based on samples from the replay buffer $B^{ex}$, the critic network $Q_\omega$ is updated based on the one-step Bellman residual $\mathcal{L}_Q(\omega)$. The policy function $\pi_\theta^{ex}$ of the executive agent is updated by maximizing the objective $\mathcal{L}_{\pi^{ex}}(\theta)$, which is constructed based on the trained critic $Q_\omega$.

## 4. Experimental results

*4.1 Data Description and Experiment Setting*

  In our research, the proposed learning system performs portfolio optimization by learning the long/short strategy. Hence, our learning system is tested on the U.S. stock market. The U.S. stock market permits short selling and is the largest and most influential stock market in the world. Hence, the U.S. stock market is the optimal market for constructing training environments and conducting back-testing experiments.

  To satisfy the assumptions (A1) and (A2) defined in Section 2.4, the selected stocks in our portfolio should have high liquidity. Hence, we select the constituted stocks in the Dow Jones Industrial Average (DJIA) to construct our portfolio. Each of the 30 companies in the DJIA is selected by the editors of the Wall Street Journal based on certain criteria, such as being a leader in its industry, having a good reputation, and being financially stable. Hence, the constituted stocks in the DJIA are typically highly liquid, with large market capitalizations and high trading volumes. After deleting the stocks with missing data, a total of 29 constituent stocks are identified to be used in our analysis. The stock data utilized spans from January 2018 to December 2022. The stock data is obtained from Yahoo Finance[1].

  In the course of this study, we carry out in-sample convergence analysis and evaluate the out-of-sample performance of our learning system. We carry out a total of four different experiments for this evaluation. Considering the dynamic nature of the financial markets [61], the parameters in our learning system's neural networks are updated every six months. In each experiment, the horizon of the training environments is set as three years. After training each agent in the learning system, we begin the back-test over the subsequent 120 trading days. Table 2 outlines the time horizons of the training and back-testing sets in each experiment. The number of trading days $K$ in a trading period is set as five in our research.

---
[1] http://www. finance.yahoo.com

| Experiment | Training set | Back-testing set |
|---|---|---|
| 1 | 2018.01.01 - 2020.12.31 | 120 trading days starting from 2021.01.01 |
| 2 | 2018.07.01 - 2021.06.30 | 120 trading days starting from 2021.07.01 |
| 3 | 2019.01.01 - 2021.12.31 | 120 trading days starting from 2022.01.01 |
| 4 | 2019.07.01 - 2022.06.30 | 120 trading days starting from 2022.07.01 |

**Table 2:** The time horizons of the training set and back-testing set in different experiments.

### *4.2 Significant Indices Tracking in the Training Process*

To demonstrate that the policy function of the DRL agent is learned through sufficient training in the environment, we constructed a tracking module to monitor the performance of the learning system in the training process. Specifically, the module is designed to track the numerical change trajectories of some significant indices regarding the accumulated return and accumulated reward in the training environment. This approach can also ensure that our HDRL training algorithm can effectively train our learning system to maximize the accumulated value of the reward and return in the training environment.

Through our experiments, we demonstrate that the design of the auxiliary agent and executive agent can help our DRL agent effectively explore the optimal policy by focusing on exploring action spaces with positive returns and lower variance. In this way, the learning system can address the issue of sparsity in positive reward and the curse of dimensionality in the exploration of the optimal policy. Hence, we track the number of days that our learning system can obtain positive returns and positive rewards in the training environment. Except for the indices tracked in the training environment, we also track the numerical change trajectories of the training objective functions used in the DRL algorithm. By tracking the numerical change trajectories of the training objective functions within the DRL algorithm, we demonstrate that, by applying the HDRL algorithm, the values of these objective functions can converge during the application of the actor-critic algorithm in the training process.

The tracked indices in the training process include two parts: the training objectives in the actor-critic algorithm and the performance indices of the learning system in the training environment. The training objectives in the actor-critic algorithm are defined in Equation (18) and (19). In the training environment, we track the numerical change trajectory of the following indices for evaluation:

- *Accumulated return* $\left(\text{AR}^{(\text{tr})}\right)$ is the sum of the logarithm return that the learning system obtained in the training environment:

$$\text{AR}^{(\text{tr})} = \log_2\left(\frac{v_{T_f^{(\text{tr})}}}{v_0}\right).$$

The index describes the profitability of our learning system in the training environment.

- *Accumulated variance* $\left(\text{AV}^{(\text{tr})}\right)$ is the sum of the variance $V_t$ that the learning system obtained in the training environment:

$$\text{AV}^{(\text{tr})} = \sigma \sum_{i=1}^{T_f^{(\text{tr})}} V_t,$$

where $\sigma$ is the amplification coefficient enabling researchers to have an intuitive sense of the numerical value. Here, the amplification coefficient $\sigma$ is calculated as:

$$\sigma = K\lambda_1,$$

where $K$ is the number of trading days in a trading period, and $\lambda_1$ is the risk aversion parameter defined in Equation (15).

- *Accumulated reward* $\left(\text{ARD}^{(\text{tr})}\right)$ is the sum of the reward $r_t^e$ that the learning system obtained in the training environment:

$$\text{ARD}^{(\text{tr})} = \sum_{i=1}^{T_f^{(\text{tr})}} r_t^{ex}.$$

Since the reward $r_t^e$ is defined in Equation (17) as an index of risk-adjusted return, the accumulated reward is an index that can describe the return and risk simultaneously.

- *Number of periods with positive return* $\left(\text{NP}^{(\text{tr})}\right)$ is the number of trading periods where the

learning system obtains the positive logarithm return in the environment in the training set:

$$\text{NP}^{(\text{tr})} = \sum_{t=1}^{T_f^{(tr)}} I(\xi_t \geq 0),$$

where $I(\cdot)$ is the indicator function. Hence, the value of $I(\xi_t)$ reflects whether the learning system obtains the positive logarithm return in the $t^{\text{th}}$ trading period. When the value of the index $\text{NP}^{(\text{tr})}$ is kept at the maximum score in the training process, we can prove that the learning system can avoid exploring the action space with negative returns.

- ***Number of periods with positive rewards*** $(\text{NPR}^{(\text{tr})})$ is the number of trading periods where the learning system obtains the positive reward in the training set environment:

$$\text{NPR}^{(\text{tr})} = \sum_{t=1}^{T_f^{(tr)}} I(r_t^{ex} \geq 0),$$

where the value of $I(r_t^e)$ reflects whether or not the reward $r_t^e$ obtained in the $t^{\text{th}}$ trading period is positive. Hence, the index $\text{NPR}^{(\text{tr})}$ gives the performance of our learning system in terms of the reward function value $r_t^{ex}$ specified in Equation (17). Suppose the value of the index increases in the training process. This means our hierarchical training algorithm can train our learning system to obtain more positive rewards in the training environment.

*4.3 In-sample Performance in the Training Process*

By tracking the above indices, we can analyze the performance of our learning system based on $\text{ARD}^{(\text{tr})}$, $\text{AR}^{(\text{tr})}$, $\text{AV}^{(\text{tr})}$, $\text{NP}^{(\text{tr})}$ and $\text{NPR}^{(\text{tr})}$. Furthermore, we carry out a convergence analysis of the training objective function when applying the DDPG algorithm. From the numerical change trajectories of the actor and critic networks' training objective function in subplot (b), the objective functions both ultimately achieve stable convergence in the training process. As given in subplot (a) of Figure 5, we can find that the value of the $\text{ARD}^{(\text{tr})}$ obtained by our learning system is improved in the training process. Furthermore, the numerical change trajectory of the $\text{NPR}^{(\text{tr})}$ given in subplot (d) indicates that our learning system can obtain more positive rewards in the training process. In subplots (a) and (b) of Figure 5, we find that, under the assistance of the auxiliary agent (BDA model of Sun et al. (2024)), our learning system can focus on exploring the optimal policy in the action space with positive return. Hence, the learning system further improves the risk control ability while maintaining outstanding profitability in the training process. It indicates that the critic network can propagate the right policy gradient direction to the policy network in maximizing the rewards obtained in the environment. Hence, our hierarchical training algorithm can achieve sufficient training of the critic network in the environment despite with the curse of dimensionality and sparsity in positive reward.

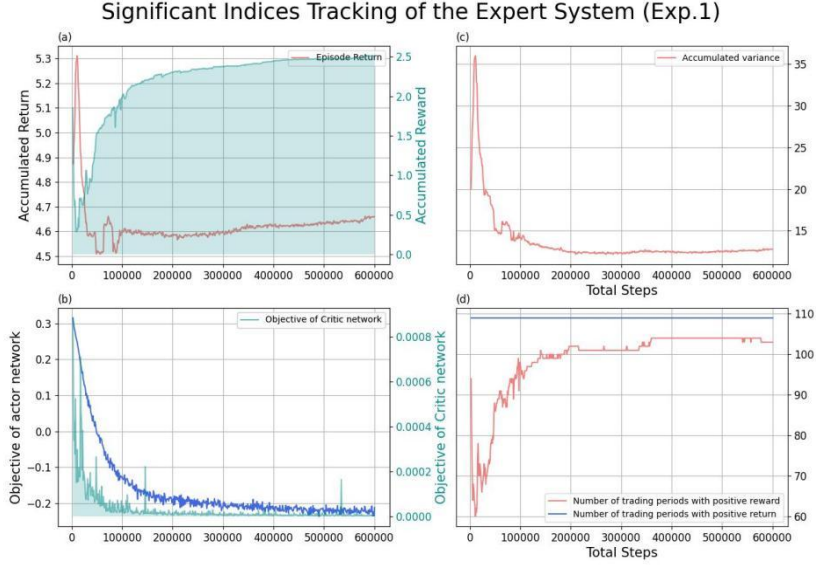

**Figure 5.** Significant indices tracking of the learning system in the training process. In the training process, we track the numerical change trajectories of some significant indices of the learning system. The tracked indices include accumulated return ($AR^{(tr)}$), accumulated reward ($ARD^{(tr)}$), accumulated variance ($AV^{(tr)}$), number of periods with positive rewards ($NPR^{(tr)}$), and number of periods with positive return ($NP^{(tr)}$), and the objective function of the actor and critic network in the DDPG algorithm. There are four subplots in the figure. Subplot (a) describes the numerical change trajectory of the $AR^{(tr)}$ and $ARD^{(tr)}$ that our learning system can obtain in the training environment. Subplot (b) describes the numerical change trajectories of the actor and critic networks' objective function in the DDPG algorithm. Subplot (c) reflects the numerical change trajectory of the $AV^{(tr)}$, and subplot (d) reflects the numerical change trajectories of the learning system's performance in the NP and NPR in the training process.

### 4.4 Back-test Performance Measure

To measure the performance of the learning system in terms of profitability and return per unit risk, we use several performance metrics to evaluate our learning system and the comparative strategies in each back-test experiment. Specifically, accumulated return (AR) and daily return (DR) are used to describe the profitability of the learning system in different back-testing horizons. Standard deviation (Std) and the lower partial standard deviation (LStd) describe the learning system's risk control ability. To consider the risk control ability and profitability simultaneously, two risk-adjusted return metrics, the Sharpe ratio (SR) and Sortino ratio (STR), are adopted to describe the return earned by considering per unit of risk.

- **Accumulated return** (AR) is the sum of daily logarithm return for the total $KT_f^{(ba)}$ trading days in the investment horizon, which is defined as:

$$AR = \sum_{t=1}^{T_f^{(ba)}} \sum_{k=1}^{K} \varpi_{t_k}, \qquad (20)$$

where $\varpi_{t_k}$ is the daily logarithm return defined in Equation (8), $T_f^{(ba)}$ is the number of trading periods in a back-test experiment, and $K$ is the number of trading days in a trading period.

- **Daily return** (DR) is the average of the daily logarithm return for all the trading days in the investment horizon, which is defined as:

$$DR = \frac{1}{KT_f} \sum_{t=1}^{T_f^{(ba)}} \sum_{k=1}^{K} \varpi_{t_k}, \qquad (21)$$

where $\varpi_{t_k}$, $T_f^{(ba)}$, and $K$ are as defined in Equation (20).

- **Standard deviation** (Std) describes the risk of the strategy, which is defined as

$$\mathrm{Std} = \sqrt{\frac{1}{KT_f^{(ba)}}\sum_{t=1}^{T_f^{(ba)}}\sum_{k=1}^{K}\left(\varpi_{t_k} - \mathrm{DR}\right)^2}, \quad (22)$$

where DR is the daily return defined in Equation (21). $\varpi_{t_k}$, $T_f^{(ba)}$, and $K$ are as defined in Equation (20).

- **Sharpe ratio** (SR) [55] is a risk-adjusted return metric to describe the excess return in a unit of risk in the whole investment horizon, which is defined as

$$\mathrm{SR} = \frac{\mathrm{DR} - r_f}{\mathrm{Std}},$$

where $r_f$ is the daily risk-free logarithm return, DR is the daily logarithm return defined in Equation (21), and Std is the standard deviation defined in Equation (22). In the time horizon of the experiments, there are substantial fluctuations in the US Federal Funds Rate, which is representative of the risk-free rate in the American financial market. In some time horizons, the US Federal Funds Rate is close to zero. Hence, in our experiments, the risk-free return $r_f$ is set as zero.

- **Lower partial standard deviation** (LStd) measures the downside risk of the strategy. It is proposed by Rollinger and Hoffman [56] based on the view that the volatility caused by the positive return could be viewed as a risk. Hence, the lower partial standard deviation (LStd) is defined as the standard deviation of the daily logarithm return $\varpi_{t_k}$ when it falls below the MAR:

$$\mathrm{LStd} = \sqrt{\frac{1}{N_L - 1}\sum_{t=1}^{T_f^{(ba)}}\sum_{k=1}^{5}(\min(\varpi_{t_k}, \mathrm{MAR}) - \mathrm{MAR})^2}, \quad (23)$$

where MAR is the minimum acceptable return, and $N_L$ is the number of samples below the minimum acceptable return. In our experiments, the minimum acceptable return (MAR) is set equal to the risk-free return. In Equation (23), $\varpi_{t_k}$, $T_f^{(ba)}$, and $K$ are as defined in Equation (20).

- **Sortino ratio** (STR) [56] is a risk-adjusted return metric which is the ratio between the value of the daily logarithm return (DR) excess the minimum acceptable return and the lower partial standard deviation (LStd):

$$\mathrm{STR} = \frac{\mathrm{DR} - \mathrm{MAR}}{\mathrm{LStd}},$$

where DR is the daily return defined in Equation (21), and LStd is the lower partial standard deviation defined in Equation (23).

### 4.5 Alternative Portfolio Strategies in the Back-tests

The current study presents the learning system based on the HDRL algorithm for portfolio management. To demonstrate the superiority of our learning system, we compare its out-of-sample performance in profitability and return per unit risk against a number of well-known portfolio choice strategies proposed by scholars in recent years. The portfolio choice strategies for comparison are divided into three distinct categories based on their underlying principles. These three underlying principles respectively are:

- Capital growth theory [4]
- Markowitz's mean-variance theory [1]
- Machine learning algorithms

In our research, we adopt fifteen strategies based on different capital growth theories, two different strategies based on Markowitz's mean-variance approach, and ten strategies based on the machine learning algorithm as the comparative strategies. The comparative strategies based on the machine learning algorithms are divided into two categories: those based on the deep learning (DL) algorithms and those based on the DRL algorithms. In the comparative strategies based on the DL algorithm, the portfolio choice strategies are constructed based on the method of Duan et al. [62]. Here, we adopt TopK-Drop strategies[2] to determine the portfolio weights based on the prediction of the DL models. In the comparative strategies based on the DRL algorithm, we adopt five DRL comparative strategies from the FinRL, which is an open-source library for financial reinforcement learning. In the framework of FinRL, the developers [63] construct a fill pipeline for the application of DRL algorithms in finance, which contains state-of-the-art DRL algorithms that are adapted to finance with fine-tuned

---

[2] https://qlib.readthedocs.io/en/latest/component/strategy.html

hyperparameters. A detailed description of these strategies is given in Table 3.

| Categories | Classifications | Algorithm |
|---|---|---|
| Strategies based on Capital Growth Theory | Baseline strategies | Constant Rebalanced Portfolios (CRP) [64] |
| | | M0 (M0) [65] |
| | | Uniform Buy And Hold (UBAH) [66] |
| | Follow-the-Winner | Universal Portfolio (UP) [64, 67] |
| | | Exponentiated Gradient (EG) [68] |
| | Follow-the-Loser | Anti-Correlation (ANTICOR) [69] |
| | | Passive Aggressive Mean Reversion (PAMR) [70] |
| | | Confidence Weights Mean Reversion (CWMR) [71] |
| | | On-Line Portfolio Selection with Moving Average Reversion (OLMAR) [72] |
| | | Robust Median Reversion (RMR) [73] |
| | | Weighted Moving Average Mean Reversion (WMAMR) [74] |
| | Pattern-Matching Approaches | Nonparametric Kernel-Based Log Optimal Strategy (BK) [75] |
| | | Correlation-driven Nonparametric learning (CORN) [76] |
| | Meta-Learning Algorithm | Online Newton Step (ONS) [77] |
| Strategies based on Mean-Variance Theory | | Jorion Bayes Stein procedure (JB) [78] |
| | | Kan and Zhou's three funds (KZTF) [79] |
| Strategies based on machine learning algorithms | DRL strategies | EIIE [20] |
| | | DDPG-FinRL [63] |
| | | SAC-FinRL [63] |
| | | PPO-FinRL [63] |
| | | TD3-FinRL [63] |
| | | A2C-FinRL [63] |
| | DL strategies | Dlinear [80] |
| | | Autoformer [81] |
| | | Informer [82] |
| | | PatchTST [83] |

**Table 3:** A detailed description of the comparative strategies and their underlying principles.

### 4.6 Out-of-Sample Performance in the Back-tests

The experimental results of our learning system and the comparative strategies are provided in Tables 3-6. In order to provide a more intuitive description of the experimental results, the performance of our learning system and the comparative strategies are visualized. Figures 6-13 are used to visually describe the experimental results. Figures 6, 8, 10, and 12 visualize the performance comparison between our learning system and the comparative strategies based on Capital growth theory and Markowitz's mean-variance theory in the four experiments. Figures 7, 9, 11, and 13 visualize the performance comparison between our learning system and the comparative strategies based on different machine learning algorithms in the four experiments.

In Figures 6-13, there are four subplots that visualize the performance of our learning system and the comparative strategies from different angles. In subplot (a) of Figures 6-13, the *x*-axis represents trading days during the back-test, and the *y*-axis represents the values of accumulative returns. In subplot (a), we observe that the trajectory of accumulative returns for each strategy as it varies over time. Subplots (b), (c), and (d), respectively, visualize the performance of our learning system and various comparative strategies according to different performance metrics. In subplot (b), the accumulative returns of our learning system and the comparative strategies during different stages of the back-testing sets (0-40 trading days, 41-80 trading days, 81-120 trading days) are reflected through bar charts of different colours. In subplot (c) and subplot (d), the *y*-axis represents DR, while the *x*-axis represents standard deviation and low partial standard deviation, respectively. In subplot (c) and subplot (d), each strategy is plotted on the graph according to their achieved DR, standard deviation, and low partial standard deviation. Hence, the slopes of the lines connecting these points to the risk-free return point provide a visual indication of the Sharpe ratio and Sortino ratio of each strategy. In subplots (c) and (d), a higher slope indicates that the strategy can achieve a higher Sharpe ratio or Sortino ratio.

Based on the subplot (a) in Figures 6, 8, 10, and 12, it can be observed that our learning system consistently outperforms the comparative strategies based on Capital growth theory or Markowitz's mean-variance theory in terms of accumulative returns in all four experiments. In addition, based on the subplot (a) in Figures 7, 9, 11, and 13, it can be observed that our learning system consistently outperforms the comparative strategies based on different machine learning algorithms in terms of accumulative returns in all four experiments. This suggests that our learning system exhibits a significant advantage in profitability compared to these comparative strategies. From subplots (b) in Figures 6, 8, 10, and 12, it is evident that our learning system consistently achieves positive returns across various stages of each back-testing set. This suggests that the profits generated by the learning system in an out-of-sample environment are robust. Drawing upon Markowitz's foundational perspective [2], portfolio optimization entails minimizing the risk per unit return or maximizing the return per unit risk. In terms of the excess return per unit risk, the Sharpe ratio and Sortino ratio are the most widely applied metrics. The subplots (c) and (d) in Figures 6, 8, 10, and 12 show that our learning system can significantly outperform the comparative strategies based on Capital growth theory or Markowitz's mean-variance theory in terms of the Sharpe ratio and Sortino ratio. Furthermore, subplots (c) and (d) in Figures 7, 9, 11, and 13 show that our learning system also significantly outperforms the comparative strategies based on different machine learning algorithms in terms of the Sharpe ratio and Sortino ratio.

The specific data of the accumulated return and return per unit risk in each experiment are given in Tables 3-6. In terms of accumulated return, except for experiment 4, the profitability of the EIIE strategy proposed by Jiang et al. [20] is slightly lower than our learning system. In other experiments, our learning system can outperform all the comparative strategies in terms of accumulative returns by at least 40.4%. Furthermore, in terms of return per unit risk, the Sharpe ratio and Sortino ratio achieved by our learning system are at least 6.3% higher than all the comparative strategies in all experiments.

These findings show that our learning system can simultaneously maximize the portfolio's accumulated return and the excess return per unit risk. The outstanding performance of our learning system in profitability and return per unit risk indicates that our learning system framework can exhibit excellent generalization ability in the back-testing experiments.

| Strategies | Performance Metrics | | | | | |
|---|---|---|---|---|---|---|
| | AR | DR | Std | SR | LStd | STR |
| **Our learning system** | **0.445735** | **0.003714** | **0.0160** | **0.2318** | **0.0148** | **0.2506** |
| BK | 0.018564 | 0.000155 | 0.0198 | 0.0078 | 0.0221 | 0.0070 |
| CRP | 0.199856 | 0.001665 | 0.0109 | 0.1526 | 0.0109 | 0.1525 |
| ONS | 0.207870 | 0.001732 | 0.0107 | 0.1624 | 0.0104 | 0.1662 |
| OLMAR | -0.059381 | -0.000495 | 0.0240 | -0.0206 | 0.0249 | -0.0199 |
| UP | 0.199661 | 0.001664 | 0.0109 | 0.1524 | 0.0109 | 0.1523 |
| Anticor | 0.173606 | 0.001447 | 0.0132 | 0.1093 | 0.0135 | 0.1069 |
| PAMR | -0.227378 | -0.001895 | 0.0225 | -0.0842 | 0.0237 | -0.0798 |
| CORN | -0.066569 | -0.000555 | 0.0220 | -0.0253 | 0.0247 | -0.0225 |
| M0 | 0.186210 | 0.001552 | 0.0125 | 0.1242 | 0.0132 | 0.1175 |
| RMR | 0.073321 | 0.000611 | 0.0241 | 0.0254 | 0.0253 | 0.0241 |
| CWMR | -0.256553 | -0.002138 | 0.0232 | -0.0920 | 0.0244 | -0.0876 |
| EG | 0.185674 | 0.001547 | 0.0113 | 0.1367 | 0.0118 | 0.1315 |
| UBAH | 0.198803 | 0.001657 | 0.0120 | 0.1385 | 0.0123 | 0.1347 |
| WMAMR | -0.072130 | -0.000601 | 0.0201 | -0.0299 | 0.0183 | -0.0329 |
| JB | -0.100375 | -0.000836 | 0.0040 | -0.2073 | 0.0044 | -0.1883 |
| KZTF | -0.098614 | -0.000822 | 0.0039 | -0.2099 | 0.0044 | -0.1855 |
| FinRL-DDPG | 0.187726 | 0.001564 | 0.0118 | 0.1326 | 0.0118 | 0.1328 |
| FinRL-A2C | 0.066760 | 0.000556 | 0.0048 | 0.1156 | 0.0042 | 0.1322 |
| FinRL-PPO | 0.062915 | 0.000524 | 0.0046 | 0.1141 | 0.0041 | 0.1279 |
| FinRL-SAC | 0.197510 | 0.001646 | 0.0109 | 0.1513 | 0.0103 | 0.1605 |
| FinRL-TD3 | 0.189876 | 0.001582 | 0.0151 | 0.1049 | 0.0150 | 0.1055 |
| EIIE | 0.273841 | 0.002282 | 0.0209 | 0.1091 | 0.0196 | 0.1166 |
| Dlinear | 0.269447 | 0.002245 | 0.0141 | 0.1589 | 0.0136 | 0.1653 |
| **Autoformer** | 0.186644 | 0.001555 | 0.0158 | 0.0984 | 0.0159 | 0.0980 |
| **PatchTST** | 0.219937 | 0.001833 | 0.0136 | 0.1351 | 0.0138 | 0.1331 |
| **Informer** | 0.186645 | 0.001555 | 0.0158 | 0.0984 | 0.0159 | 0.0980 |

**Table 3:** Empirical results in experiment 1. The table presents the quantitative results of the indices defined in Section 4.4 for our learning system and several comparative strategies in the back-test experiment 1. The superior results for the return and return per unit risk metrics are emphasized in bold.

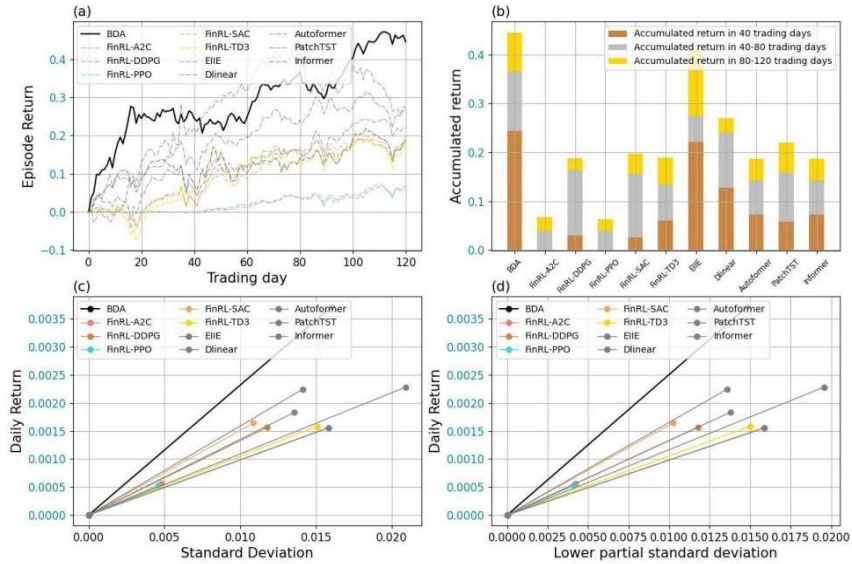

**Figure 6.** The performance of different portfolio choice algorithms in experiment 1. The figure depicts the performance comparison of our HDRL-based learning system and various strategies based on the Capital growth theory or Markowitz's mean-variance theory. The figure is the visualization of the empirical results reflected in Table 3.

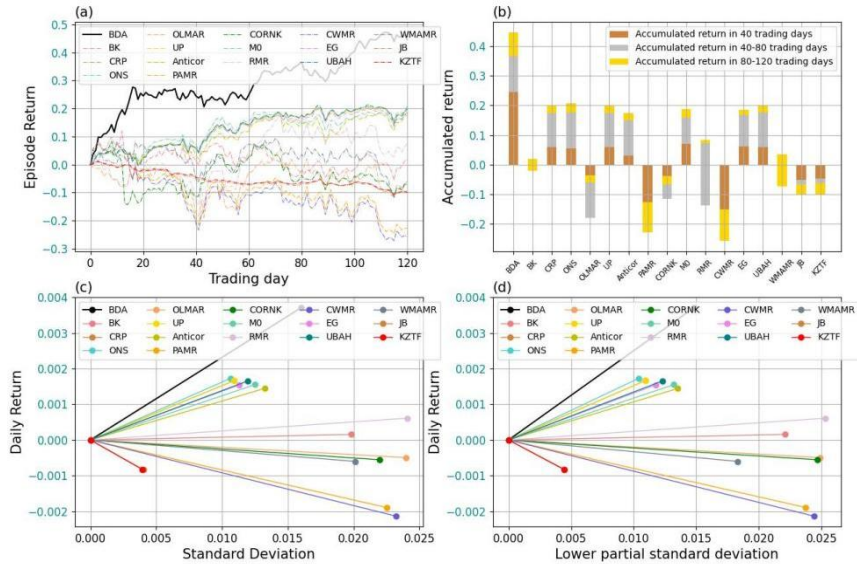

**Figure 7.** The performance of different portfolio choice strategies in experiment 1. The figure depicts the performance comparison of our HDRL-based learning system and various strategies based on different machine-learning algorithms. The figure is the visualization of the empirical results reflected in Table 3.

| Strategies | Performance Metrics |
|---|---|

|  | AR | DR | Std | SR | LStd | STR |
|---|---|---|---|---|---|---|
| **Our learning system** | **0.372162** | **0.003101** | 0.0256 | **0.1213** | 0.0268 | **0.1159** |
| **BK** | -0.064097 | -0.000534 | 0.0139 | -0.0385 | 0.0160 | -0.0333 |
| **CRP** | 0.043106 | 0.000359 | 0.0110 | 0.0326 | 0.0116 | 0.0311 |
| **ONS** | 0.048112 | 0.000401 | 0.0113 | 0.0356 | 0.0117 | 0.0341 |
| **OLMAR** | -0.142528 | -0.001188 | 0.0224 | -0.0531 | 0.0210 | -0.0567 |
| **UP** | 0.043305 | 0.000361 | 0.0110 | 0.0328 | 0.0116 | 0.0312 |
| **Anticor** | 0.098246 | 0.000819 | 0.0137 | 0.0597 | 0.0119 | 0.0686 |
| **PAMR** | -0.247310 | -0.002061 | 0.0211 | -0.0976 | 0.0223 | -0.0923 |
| **CORN** | -0.033465 | -0.000279 | 0.0183 | -0.0153 | 0.0207 | -0.0135 |
| **M0** | 0.022049 | 0.000184 | 0.0125 | 0.0147 | 0.0134 | 0.0137 |
| **RMR** | -0.032823 | -0.000274 | 0.0228 | -0.0120 | 0.0221 | -0.0124 |
| **CWMR** | -0.257192 | -0.002143 | 0.0218 | -0.0984 | 0.0228 | -0.0940 |
| **EG** | 0.036990 | 0.000308 | 0.0108 | 0.0284 | 0.0121 | 0.0256 |
| **UBAH** | 0.035635 | 0.000297 | 0.0110 | 0.0270 | 0.0117 | 0.0254 |
| **WMAMR** | -0.056237 | -0.000469 | 0.0218 | -0.0215 | 0.0207 | -0.0227 |
| **JB** | 0.033133 | 0.000276 | 0.0085 | 0.0323 | 0.0096 | 0.0287 |
| **KZTF** | 0.044739 | 0.000373 | 0.0076 | 0.0493 | 0.0080 | 0.0467 |
| **FinRL-DDPG** | 0.044474 | 0.000371 | 0.0114 | 0.0324 | 0.0128 | 0.0290 |
| **FinRL-A2C** | 0.054653 | 0.000455 | 0.0059 | 0.0775 | 0.0056 | 0.0808 |
| **FinRL-PPO** | 0.007362 | 0.000061 | 0.0066 | 0.0094 | 0.0068 | 0.0090 |
| **FinRL-SAC** | 0.046984 | 0.000392 | 0.0119 | 0.0328 | 0.0126 | 0.0311 |
| **FinRL-TD3** | -0.012638 | -0.000105 | 0.0122 | -0.0086 | 0.0124 | -0.0085 |
| **EIIE** | 0.053187 | 0.000443 | 0.0142 | 0.0312 | 0.0139 | 0.0320 |
| **Dlinear** | -0.061676 | -0.000514 | 0.0139 | -0.0370 | 0.0149 | -0.0345 |
| **Autoformer** | -0.037963 | -0.000316 | 0.0138 | -0.0229 | 0.0135 | -0.0234 |
| **PatchTST** | 0.039936 | 0.000333 | 0.0144 | 0.0231 | 0.0150 | 0.0221 |
| **Informer** | -0.023890 | -0.000199 | 0.0139 | -0.0144 | 0.0136 | -0.0146 |

**Table 4:** Empirical results in experiment 2. The indices displayed in this table are in agreement with those outlined in Table 3. The superior results for the return and return per unit risk metrics are emphasized in bold.

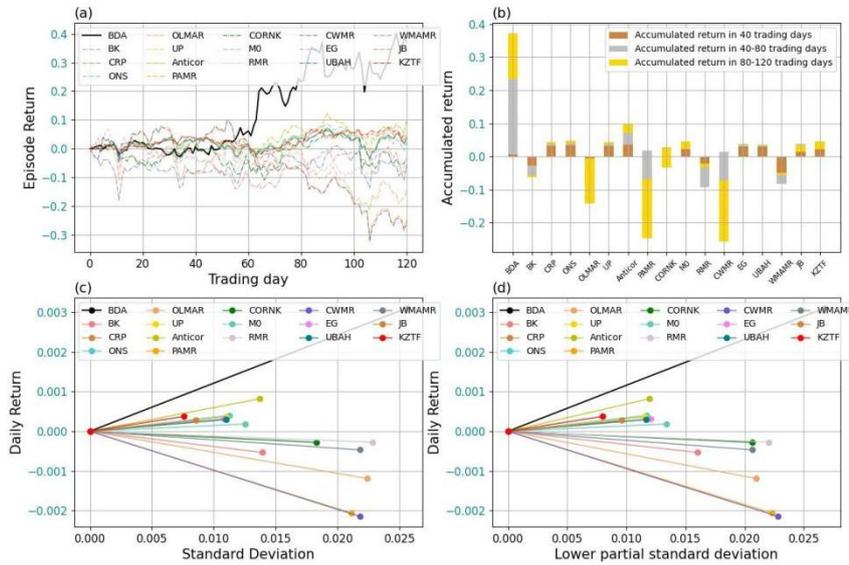

**Figure 8.** The performance of different portfolio choice strategies in experiment 2. The strategies and indices visualized in the figure are consistent with those in Figure 6.

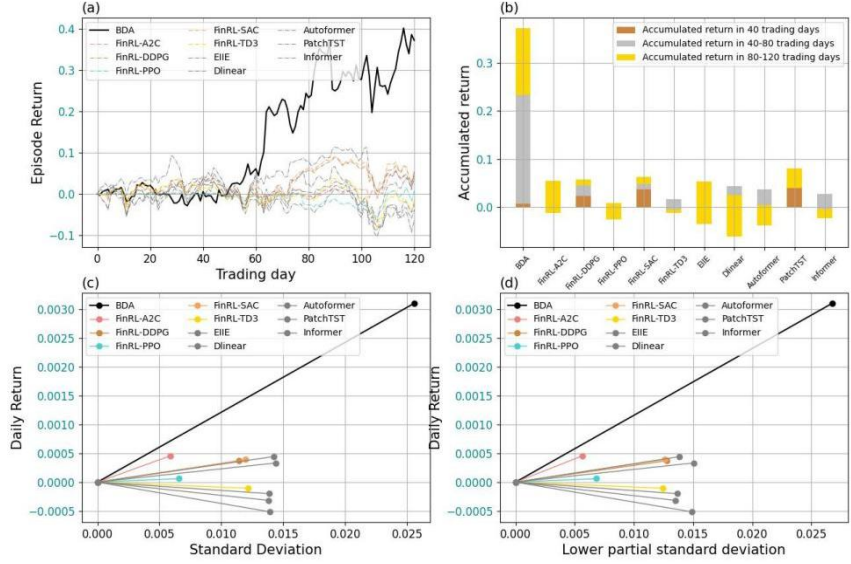

**Figure 9.** The performance of different portfolio choice strategies in experiment 2. The strategies and indices visualized in the figure are consistent with those in Figure 7. The superior results for the return and return per unit risk metrics are emphasized in bold.

| Strategies | Performance Metrics | | | | | |
| --- | --- | --- | --- | --- | --- | --- |
| | AR | DR | Std | SR | LStd | STR |
| **Our learning system** | **0.511425** | **0.004262** | 0.0431 | **0.0989** | 0.0384 | **0.1110** |
| BK | -0.128192 | -0.001068 | 0.0225 | -0.0474 | 0.0263 | -0.0406 |
| CRP | -0.184698 | -0.001539 | 0.0180 | -0.0853 | 0.0191 | -0.0806 |
| ONS | -0.192840 | -0.001607 | 0.0192 | -0.0838 | 0.0197 | -0.0815 |
| OLMAR | -0.505747 | -0.004215 | 0.0344 | -0.1225 | 0.0361 | -0.1168 |
| UP | -0.184581 | -0.001538 | 0.0180 | -0.0854 | 0.0191 | -0.0806 |
| Anticor | -0.165465 | -0.001379 | 0.0254 | -0.0543 | 0.0257 | -0.0536 |
| PAMR | -0.615435 | -0.005129 | 0.0348 | -0.1475 | 0.0382 | -0.1342 |
| CORN | -0.151086 | -0.001259 | 0.0238 | -0.0528 | 0.0269 | -0.0468 |
| M0 | -0.236546 | -0.001971 | 0.0197 | -0.1003 | 0.0212 | -0.0930 |
| RMR | -0.678905 | -0.005658 | 0.0347 | -0.1630 | 0.0377 | -0.1500 |
| CWMR | -0.626532 | -0.005221 | 0.0357 | -0.1461 | 0.0388 | -0.1346 |
| EG | -0.173681 | -0.001447 | 0.0167 | -0.0867 | 0.0177 | -0.0817 |
| UBAH | -0.179335 | -0.001494 | 0.0161 | -0.0928 | 0.0173 | -0.0865 |
| WMAMR | -0.192369 | -0.001603 | 0.0329 | -0.0487 | 0.0348 | -0.0461 |
| JB | -0.129447 | -0.001079 | 0.0096 | -0.1119 | 0.0112 | -0.0965 |
| KZTF | -0.096956 | -0.000808 | 0.0085 | -0.0956 | 0.0097 | -0.0837 |
| FinRL-DDPG | -0.278260 | -0.002319 | 0.0210 | -0.1102 | 0.0218 | -0.1064 |
| FinRL-A2C | -0.006823 | -0.000057 | 0.0102 | -0.0056 | 0.0103 | -0.0055 |
| FinRL-PPO | -0.046191 | -0.000385 | 0.0132 | -0.0291 | 0.0140 | -0.0276 |
| FinRL-SAC | -0.214470 | -0.001787 | 0.0216 | -0.0827 | 0.0225 | -0.0793 |
| FinRL-TD3 | -0.172236 | -0.001435 | 0.0189 | -0.0759 | 0.0194 | -0.0738 |
| EIIE | -0.376169 | -0.003135 | 0.0346 | -0.0906 | 0.0354 | -0.0886 |
| Dlinear | -0.256898 | -0.002141 | 0.0227 | -0.0944 | 0.0235 | -0.0909 |
| Autoformer | -0.327203 | -0.002727 | 0.0199 | -0.1368 | 0.0206 | -0.1322 |
| PatchTST | -0.180488 | -0.001504 | 0.0199 | -0.0757 | 0.0212 | -0.0708 |
| Informer | -0.327203 | -0.002727 | 0.0199 | -0.1368 | 0.0206 | -0.1322 |

**Table 5:** Empirical results in experiment 3. The indices displayed in this table are in agreement with those outlined in Table 3.

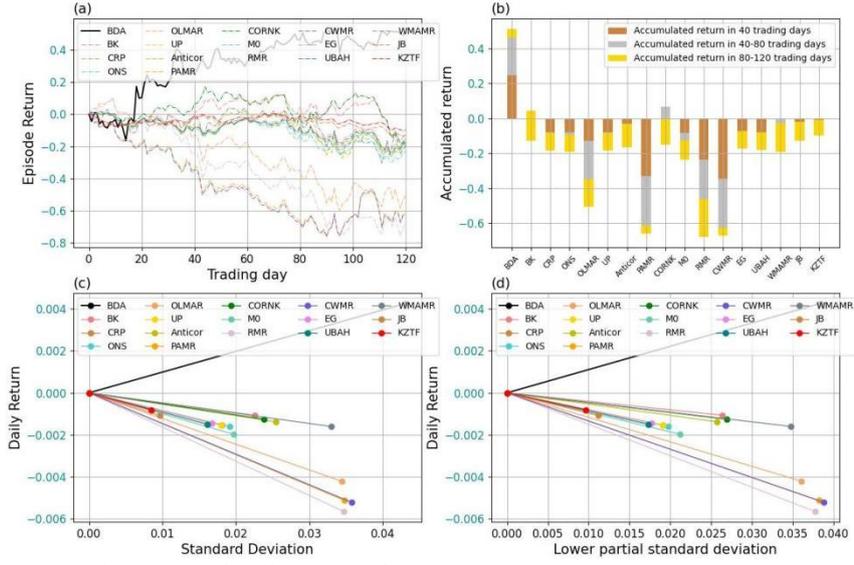

**Figure 10.** The performance of different portfolio choice strategies in experiment 3. The strategies and indices visualized in the figure are consistent with those in Figure 6.

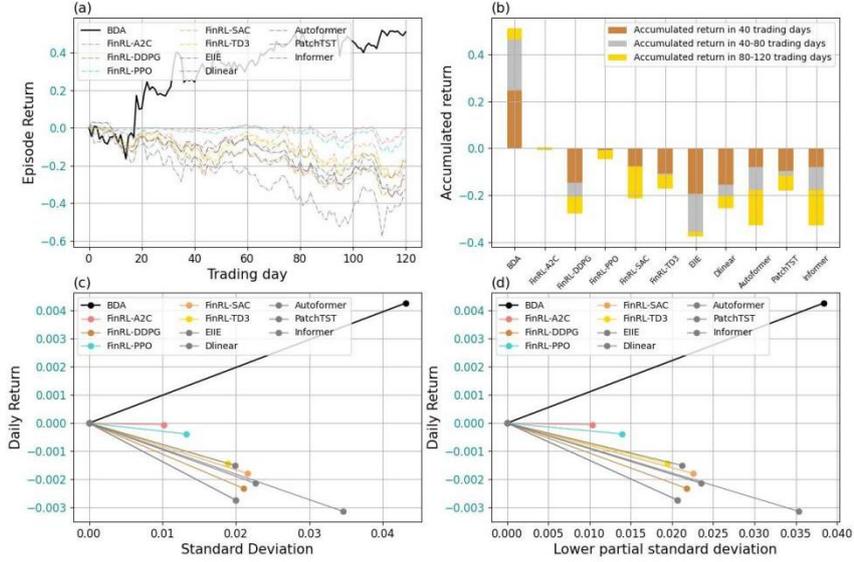

**Figure 11.** The performance of different portfolio choice strategies in experiment 3. The strategies and indices visualized in the figure are consistent with those in Figure 7.

| Strategies | Performance Metrics | | | | | |
|---|---|---|---|---|---|---|
| | AR | DR | Std | SR | LStd | STR |
| **Our learning system** | **0.243706** | **0.002031** | 0.0217 | **0.0936** | 0.0228 | **0.0892** |
| **BK** | 0.078252 | 0.000652 | 0.0209 | 0.0312 | 0.0213 | 0.0306 |
| **CRP** | 0.099332 | 0.000828 | 0.0185 | 0.0448 | 0.0178 | 0.0465 |
| **ONS** | 0.096636 | 0.000805 | 0.0186 | 0.0433 | 0.0178 | 0.0453 |
| **OLMAR** | 0.165243 | 0.001377 | 0.0353 | 0.0390 | 0.0318 | 0.0434 |
| **UP** | 0.099923 | 0.000833 | 0.0185 | 0.0451 | 0.0178 | 0.0467 |
| **Anticor** | 0.113945 | 0.000950 | 0.0240 | 0.0395 | 0.0217 | 0.0437 |
| **PAMR** | -0.592665 | -0.004939 | 0.0348 | -0.1421 | 0.0372 | -0.1326 |
| **CORN** | 0.166555 | 0.001388 | 0.0226 | 0.0614 | 0.0225 | 0.0617 |
| **M0** | 0.100311 | 0.000836 | 0.0198 | 0.0422 | 0.0185 | 0.0452 |
| **RMR** | 0.013514 | 0.000113 | 0.0379 | 0.0030 | 0.0373 | 0.0030 |
| **CWMR** | -0.612289 | -0.005102 | 0.0353 | -0.1446 | 0.0376 | -0.1357 |
| **EG** | 0.097508 | 0.000813 | 0.0184 | 0.0441 | 0.0180 | 0.0452 |
| **UBAH** | 0.086772 | 0.000723 | 0.0186 | 0.0388 | 0.0184 | 0.0393 |
| **WMAMR** | 0.091803 | 0.000765 | 0.0327 | 0.0234 | 0.0341 | 0.0224 |
| **JB** | -0.018912 | -0.000158 | 0.0076 | -0.0206 | 0.0088 | -0.0179 |
| **KZTF** | -0.008455 | -0.000070 | 0.0067 | -0.0105 | 0.0076 | -0.0092 |
| **FinRL-DDPG** | 0.083452 | 0.000695 | 0.0208 | 0.0334 | 0.0187 | 0.0373 |
| **FinRL-A2C** | 0.031667 | 0.000264 | 0.0095 | 0.0277 | 0.0089 | 0.0296 |
| **FinRL-PPO** | 0.038682 | 0.000322 | 0.0125 | 0.0258 | 0.0115 | 0.0279 |
| **FinRL-SAC** | 0.096154 | 0.000801 | 0.0192 | 0.0417 | 0.0177 | 0.0452 |
| **FinRL-TD3** | 0.066808 | 0.000557 | 0.0144 | 0.0386 | 0.0135 | 0.0413 |
| **EIIE** | 0.240184 | 0.002002 | 0.0255 | 0.0784 | 0.0238 | 0.0839 |
| **Dlinear** | 0.119330 | 0.000994 | 0.0224 | 0.0443 | 0.0222 | 0.0449 |
| **Autoformer** | 0.123590 | 0.001030 | 0.0169 | 0.0611 | 0.0174 | 0.0591 |
| **PatchTST** | 0.173488 | 0.001446 | 0.0203 | 0.0713 | 0.0190 | 0.0763 |
| **Informer** | 0.123590 | 0.001030 | 0.0169 | 0.0611 | 0.0174 | 0.0591 |

**Table 6:** Empirical results in experiment 4. The indices displayed in this table are in agreement with those outlined in Table 3. The superior results for the return and return per unit risk metrics are emphasized in bold.

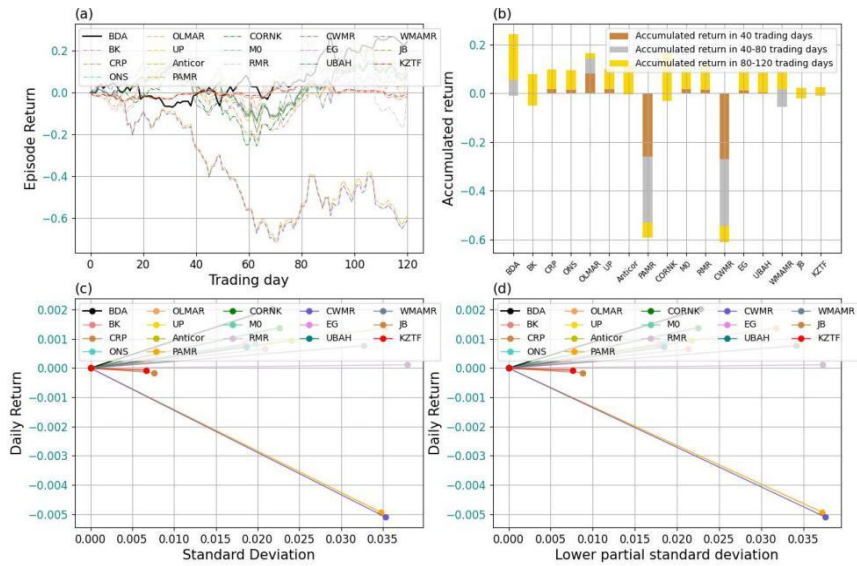

**Figure 12.** The performance of different portfolio choice strategies in experiment 4. The strategies and indices visualized in the figure are consistent with those in Figure 6.

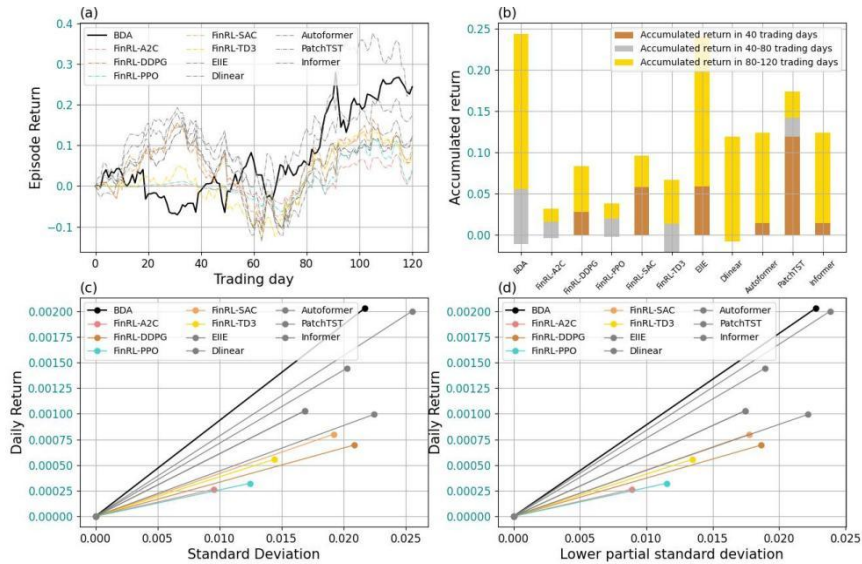

**Figure 13.** The performance of different portfolio choice strategies in experiment 4. The strategies and indices visualized in the figure are consistent with those in Figure 7.

*4.7 Ablation study*

To investigate the necessity of the application of each DRL agent in the framework of our learning system, we conduct the following ablation studies. We construct two ablation studies in the four experiments mentioned in Section 4.1, and the empirical results in the back-tests are given in Table 7.

**The necessity of the executive agent**

In the first ablation study, we compare our learning system with a variant of our learning system. In this variant, we ablate the executive agent in the learning system, and the target portfolio weights at the beginning of each trading period are solely determined by the auxiliary agent, which is the BL-based DRL agent (BDA) proposed by Sun et al. [54]. This variant is termed learning system variant 1 (LSV1) in our paper. The empirical results in the back-tests show that our learning system outperforms the variant LSV1 in Sharpe ratio and Sortino ratio by at least 9.7%. In terms of profitability, the variant LSV1 only performs better than our learning system in experiment 4. This is due to the fact that, in Experiment 4, LSV1 assumes significantly greater investment risk than our learning system. As shown in Table 7, both the Std and the LStd of LSV1 are at least 54% higher than those of our learning system. Nevertheless, in the other three experiments, our learning system still can outperform the variant LSV1. It demonstrates that the executive agent is essential in ensuring the out-of-sample generalization ability of the learning system in profitability and return per unit risk.

**The necessity of the auxiliary agent**

To prove the necessity of the auxiliary task done by the auxiliary agent in dynamic portfolio optimization, we propose another variant (LSV2). In the variant LSV2, we ablate the auxiliary agent in the framework of our learning system and replace the baseline portfolio weights determined by the auxiliary agent with the equilibrium weights. The empirical results in the back-tests show that the accumulated return of our learning system is at least seven times higher than that of the variant LSV2. In terms of return per unit risk, the Sharpe ratio and Sortino ratio of our learning system are at least 54.5% higher than that of the variant LSV2. It suggests that the auxiliary agent is essential in ensuring the generalization ability of the learning system in profitability. Without the baseline portfolio weights provided by the auxiliary agent, it is difficult for the learning system to obtain considerable accumulated returns in the back-test environments. Furthermore, the auxiliary agent also plays a crucial role in ensuring the learning system's out-of-sample generalization ability in return per unit risk.

| Exp. | Name | Performance Metrics | | | | | |
| --- | --- | --- | --- | --- | --- | --- | --- |
| | | AR | DR | Std | SR | LStd | STR |
| 1 | **Our learning system** | **0.445735** | **0.003714** | 0.0160 | **0.2318** | 0.0148 | **0.2506** |
| | LSV1 | 0.397122 | 0.003309 | 0.0237 | 0.1394 | 0.0238 | 0.1390 |
| | LSV2 | 0.050254 | 0.000419 | 0.0028 | 0.1500 | 0.0028 | 0.1472 |
| 2 | **Our learning system** | **0.372162** | **0.003101** | 0.0256 | **0.1213** | 0.0268 | **0.1159** |
| | LSV1 | 0.274825 | 0.002290 | 0.0262 | 0.0875 | 0.0241 | 0.0952 |
| | LSV2 | 0.011964 | 0.000100 | 0.0027 | 0.0365 | 0.0029 | 0.0349 |
| 3 | **Our learning system** | **0.511425** | **0.004262** | 0.0431 | **0.0989** | 0.0384 | **0.1110** |
| | LSV1 | 0.344696 | 0.002872 | 0.0347 | 0.0828 | 0.0301 | 0.0954 |
| | LSV2 | -0.043005 | -0.000358 | 0.0047 | -0.0759 | 0.0050 | -0.0721 |
| 4 | **Our learning system** | 0.243706 | 0.002031 | 0.0217 | **0.0936** | 0.0228 | **0.0892** |
| | LSV1 | **0.343216** | **0.002860** | 0.0351 | 0.0814 | 0.0352 | 0.0813 |
| | LSV2 | 0.021491 | 0.000179 | 0.0048 | 0.0370 | 0.0046 | 0.0387 |

**Table 7:** Empirical results in the ablation studies. The best results for each metric are highlighted in bold.

# 5. Conclusions and Future Work

This paper proposes a multi-agent hierarchical deep reinforcement learning framework as the learning system for portfolio optimization in consecutive trading periods. In this learning system, the auxiliary agent and the executive agent collaborate to determine the target portfolio weights. Specifically, the auxiliary agent is trained to provide the baseline portfolio weights to assist the executive agent in exploring the optimal portfolio optimization strategy. In the training process, by adopting the auxiliary agent to assist the executive agent in the determination of the target portfolio weight, we can effectively overcome the training difficulties caused by the issues of positive reward sparse and curse of dimensionality when training the executive agent using the actor-critic algorithm and deep function approximators. Hence, we can achieve sufficient training of the critic network. In the out-of-sample performance, the empirical results demonstrate that our learning system can significantly outperform the traditional portfolio optimization strategies and the strategies based on different machine learning algorithms in accumulative returns and the return per unit risk. Furthermore, the results of ablation studies also demonstrate that our learning system significantly outperforms the performance of each individual agent in terms of return per unit risk in the back-tests. It demonstrates that our HDRL training algorithm can ensure the generalization ability of the learning system's policy.

Although we achieve favourable empirical results, our learning system still has the following limitations. First, the performance of our learning system exhibits significant discrepancies in terms of profitability and risk control between its performance in the back-testing set and its performance in the training set. It reflects that the generalization ability of our learning system is insufficient. Second, the BL model used by the auxiliary agent employs a Bayesian model based on a Gaussian distribution for portfolio decision-making. However, in actual financial markets, the return distribution often exhibits heavier tails and occasionally high peaks, which cannot be accurately described by a Gaussian distribution [84]. This can result in the DRL agent failing to capture certain crucial features of the return distribution, making it difficult for the DRL agent to accurately describe the return distribution of assets in the market.

In future research, we will attempt to address these limitations. First, we will employ data augmentation techniques during the training of the DRL agents in our learning system to enhance the policy's out-of-sample generalization ability in their decision-making. Specifically, we plan to leverage data augmentation as an optimality-invariant state transformation to induce a bisimulation relation between states and their transformed (augmented) counterparts. Based on the original MDPs, it can generalize to novel MDPs with the augmented state. We aim to train the DRL agent to learn the policy, which generalizes well to generated MDPs. In this way, the issue of the poor generalization ability of our learning system can be overcome. Second, we will construct a Bayesian model based on distributions that can describe heavier tails and high peaks (i.e., Elliptical distributions [84]) as a replacement for the BL model used by the auxiliary agent in our learning system. Consequently, the learning system can better describe the return distribution and effectively capture the crucial features of the environment when determining the baseline portfolio weights in the learning system.

**Data availability**

This research uses only publicly available American stock data from Yahoo Finance platform. All datasets generated during and/or analyzed during the current study, models, or codes that support the findings of this study are available from the corresponding author upon reasonable request.

**Declarations Conflict of interest**

On behalf of all authors, the corresponding author states that there is no conflict of interest.

# Reference

[1] Markowitz , H. (1952). Portfolio selection. The Journal of Finance 7, 1, 77–91.
[2] Markovitz, H. (1959). Portfolio selection: Efficient diversification of investments. *NY: John Wiley*.
[3] Markowitz, H. M., & Todd, G. P. (2000). *Mean-variance analysis in portfolio choice and capital markets* (Vol. 66). John Wiley & Sons.
[4] Kelly, J. L. (1956). A new interpretation of information rate. *the bell system technical journal*, *35*(4), 917-926.
[5] Hakansson, N. H., & Ziemba, W. T. (1995). Capital growth theory. *Handbooks in operations research and management science*, *9*, 65-86.
[6] Li, B., & Hoi, S. C. (2014). Online portfolio selection: A survey. *ACM Computing Surveys (CSUR)*, *46*(3), 1-36.
[7] Du, J., Jin, M., Kolm, P. N., Ritter, G., Wang, Y., & Zhang, B. (2020). Deep reinforcement learning for option replication and hedging. *The Journal of Financial Data Science*, *2*(4), 44-57.
[8] Seong, N., & Nam, K. (2021). Predicting stock movements based on financial news with segmentation. *Expert Systems with Applications*, *164*, 113988.
[9] Song, Q., Liu, A., & Yang, S. Y. (2017). Stock portfolio selection using learning-to-rank algorithms with news sentiment. *Neurocomputing*, *264*, 20-28.
[10] Nam, K., & Seong, N. (2019). Financial news-based stock movement prediction using causality analysis of influence in the Korean stock market. *Decision Support Systems*, *117*, 100-112.
[11] Atsalakis, G. S., & Valavanis, K. P. (2009). Surveying stock market forecasting techniques part II: Soft computing methods. *Expert Systems with Applications, 36* (3), 5932–5941. https://doi.org/10.1016/j.ijar.2014.07.005
[12] Jing, N., Wu, Z., & Wang, H. (2021). A hybrid model integrating deep learning with investor sentiment analysis for stock price prediction. *Expert Systems with Applications*, *178*, 115019.
[13] Carta, S., Consoli, S., Podda, A. S., Recupero, D. R., & Stanciu, M. M. (2022). Statistical arbitrage powered by explainable artificial intelligence. *Expert Systems with Applications*, 206, Article 117763.
[14] Jang, J., & Seong, N. (2023). Deep reinforcement learning for stock portfolio optimization by connecting with modern portfolio theory. *Expert Systems with Applications*, *218*, 119556.
[15] Jiang, W., Liu, M., Xu, M., Chen, S., Shi, K., Liu, P., ... & Zhao, F. (2024). New reinforcement learning based on representation transfer for portfolio management. *Knowledge-Based Systems*, *293*, 111697.
[16] Kang, M., Templeton, G. F., Kwak, D. H., & Um, S. (2024). Development of an AI Framework Using Neural Process Continuous Reinforcement Learning to Optimize Highly Volatile Financial Portfolios. *Knowledge-Based Systems*, 112017.
[17] Hirshleifer, D. (2015). Behavioral finance. *Annual Review of Financial Economics*, *7*, 133-159.
[18] Hambly, B., Xu, R., & Yang, H. (2023). Recent advances in reinforcement learning in finance. *Mathematical Finance*, *33*(3), 437-503.
[19] Yang, S. (2023). Deep reinforcement learning for portfolio management. *Knowledge-Based Systems*, *278*, 110905.
[20] Jiang, Z., Xu, D., & Liang, J. (2017). A deep reinforcement learning framework for the financial portfolio management problem. *arXiv preprint arXiv:1706.10059*.
[21] Shi, S., Li, J., Li, G., & Pan, P. (2019, November). A multi-scale temporal feature aggregation convolutional neural network for portfolio management. In *Proceedings of the 28th ACM international conference on information and knowledge management* (pp. 1613-1622).
[22] Ye, Y., Pei, H., Wang, B., Chen, P. Y., Zhu, Y., Xiao, J., & Li, B. (2020, April). Reinforcement-learning based portfolio management with augmented asset movement prediction states. In *Proceedings of the AAAI conference on artificial intelligence* (Vol. 34, No. 01, pp. 1112-1119).
[23] Wang, J., Zhang, Y., Tang, K., Wu, J., & Xiong, Z. (2019, July). Alphastock: A buying-winners-and-selling-losers investment strategy using interpretable deep reinforcement attention networks. In *Proceedings of the 25th ACM SIGKDD international conference on knowledge discovery & data mining* (pp. 1900-1908).
[24] Shi, S., Li, J., Li, G., Pan, P., Chen, Q., & Sun, Q. (2022). GPM: A graph convolutional network based reinforcement learning framework for portfolio management. *Neurocomputing*, *498*, 14-27.
[25] Yu, P., Lee, J. S., Kulyatin, I., Shi, Z., & Dasgupta, S. (2019). Model-based deep reinforcement learning for dynamic portfolio optimization. *arXiv preprint arXiv:1901.08740*.
[26] Wang, Z., Huang, B., Tu, S., Zhang, K., & Xu, L. (2021, May). Deeptrader: A deep reinforcement learning approach for risk-return balanced portfolio management with market conditions embedding. In *Proceedings of the AAAI conference on artificial intelligence* (Vol. 35, No. 1, pp. 643-650).

[27] Zhang, Y., Zhao, P., Wu, Q., Li, B., Huang, J., & Tan, M. (2020). Cost-sensitive portfolio selection via deep reinforcement learning. *IEEE Transactions on Knowledge and Data Engineering*, *34*(1), 236-248.
[28] Sun, R., Jiang, Z., & Su, J. (2021, March). A deep residual shrinkage neural network-based deep reinforcement learning strategy in financial portfolio management. In *2021 IEEE 6th International Conference on Big Data Analytics (ICBDA)* (pp. 76-86). IEEE.
[29] Ren, X., Jiang, Z., & Su, J. (2021, March). The use of features to enhance the capability of deep reinforcement learning for investment portfolio management. In *2021 IEEE 6th International Conference on Big Data Analytics (ICBDA)* (pp. 44-50). IEEE.
[30] Gu, F., Jiang, Z., & Su, J. (2021, March). Application of features and neural network to enhance the performance of deep reinforcement learning in portfolio management. In *2021 IEEE 6th International Conference on Big Data Analytics (ICBDA)* (pp. 92-97). IEEE.
[31] Cui, B., Sun, R., & Su, J. (2022, March). A Novel Deep Reinforcement Learning Strategy in Financial Portfolio Management. In *2022 7th International Conference on Big Data Analytics (ICBDA)* (pp. 341-348). IEEE.
[32] Zhang, R., Ren, X., Gu, F., Stefanidis, A., Sun, R., & Su, J. (2022, October). MDAEN: Multi-Dimensional Attention-based Ensemble Network in Deep Reinforcement Learning Framework for Portfolio Management. In *2022 International Conference on Cyber-Enabled Distributed Computing and Knowledge Discovery (CyberC)* (pp. 143-151). IEEE.
[33] Yang, X., Sun, R., Ren, X., Stefanidis, A., Gu, F., & Su, J. (2022, October). Ghost Expectation Point with Deep Reinforcement Learning in Financial Portfolio Management. In *2022 International Conference on Cyber-Enabled Distributed Computing and Knowledge Discovery (CyberC)* (pp. 136-142). IEEE.
[34] Gao, R., Gu, F., Sun, R., Stefanidis, A., Ren, X., & Su, J. (2022, October). A Novel DenseNet-based Deep Reinforcement Framework for Portfolio Management. In *2022 International Conference on Cyber-Enabled Distributed Computing and Knowledge Discovery (CyberC)* (pp. 158-165). IEEE.
[35] Yang, B., Liang, T., Xiong, J., & Zhong, C. (2023). Deep reinforcement learning based on transformer and U-Net framework for stock trading. *Knowledge-Based Systems*, *262*, 110211.
[36] Du, X., Zhai, J., & Lv, K. (2016). Algorithm trading using q-learning and recurrent reinforcement learning. *positions*, *1*(1), 1-7.
[37] Park, H., Sim, M. K., & Choi, D. G. (2020). An intelligent financial portfolio trading strategy using deep Q-learning. *Expert Systems with Applications*, *158*, 113573.
[38] Pendharkar, P. C., & Cusatis, P. (2018). Trading financial indices with reinforcement learning agents. *Expert Systems with Applications*, *103*, 1-13.
[39] Lucarelli, G., & Borrotti, M. (2020). A deep Q-learning portfolio management framework for the cryptocurrency market. *Neural Computing and Applications*, *32*, 17229-17244.
[40] Gao, Z., Gao, Y., Hu, Y., Jiang, Z., & Su, J. (2020, May). Application of deep q-network in portfolio management. In *2020 5th IEEE International Conference on Big Data Analytics (ICBDA)* (pp. 268-275). IEEE.
[41] Gao, Y., Gao, Z., Hu, Y., Song, S., Jiang, Z., & Su, J. (2021, February). A Framework of Hierarchical Deep Q-Network for Portfolio Management. In *ICAART (2)* (pp. 132-140).
[42] Lillicrap, T. P., Hunt, J. J., Pritzel, A., Heess, N., Erez, T., Tassa, Y., ... & Wierstra, D. (2015). Continuous control with deep reinforcement learning. *arXiv preprint arXiv:1509.02971*.
[43] Felizardo, L. K., Paiva, F. C. L., Costa, A. H. R., & Del-Moral-Hernandez, E. (2022). Reinforcement Learning Applied to Trading Systems: A Survey. *arXiv preprint arXiv:2212.06064*.
[44] Haarnoja, T., Zhou, A., Abbeel, P., & Levine, S. (2018, July). Soft actor-critic: Off-policy maximum entropy deep reinforcement learning with a stochastic actor. In *International conference on machine learning* (pp. 1861-1870). PMLR
[45] Schulman, J., Wolski, F., Dhariwal, P., Radford, A., & Klimov, O. (2017). Proximal policy optimization algorithms. *arXiv preprint arXiv:1707.06347*.
[46] Aremu, O. O., Hyland-Wood, D., & McAree, P. R. (2020). A machine learning approach to circumventing the curse of dimensionality in discontinuous time series machine data. *Reliability Engineering & System Safety*, *195*, 106706.
[47] Ocana, J. M. C., Capobianco, R., & Nardi, D. (2023). An overview of environmental features that impact deep reinforcement learning in sparse-reward domains. *Journal of Artificial Intelligence Research*, *76*, 1181-1218.
[48] Bellman, R. (1957). *Dynamic programming*. Princeton University Press, Princeton, NJ, USA.
[49] Curran, W., Brys, T., Taylor, M., & Smart, W. (2015). Using PCA to efficiently represent state spaces. *arXiv preprint arXiv:1505.00322*.

[50] Hao, X., Mao, H., Wang, W., Yang, Y., Li, D., Zheng, Y., ... & Hao, J. (2022). Breaking the curse of dimensionality in multiagent state space: A unified agent permutation framework. *arXiv preprint arXiv:2203.05285*.
[51] Barto, A. G., & Mahadevan, S. (2003). Recent advances in hierarchical reinforcement learning. Discrete event dynamic systems, 13, 341-379.
[52] Schmidhuber, J. (2010). Formal theory of creativity, fun, and intrinsic motivation (1990–2010). *IEEE transactions on autonomous mental development*, *2*(3), 230-247.
[53] Hutsebaut-Buysse, M., Mets, K., & Latré, S. (2022). Hierarchical reinforcement learning: A survey and open research challenges. *Machine Learning and Knowledge Extraction*, *4*(1), 172-221.
[54] Sun, R., Stefanidis, A., Jiang, Z., & Su, J. (2024). Combining transformer based deep reinforcement learning with Black-Litterman model for portfolio optimization. *Neural Computing & Application*. https://doi.org/10.1007/s00521-024-09805-9
[55] Sharpe, W. F. (1966). Mutual fund performance. *The Journal of business*, *39*(1), 119-138.
[56] Rollinger, T. N., & Hoffman, S. T. (2013). Sortino: a 'sharper'ratio. *Chicago, Illinois: Red Rock Capital*.
[57] Li, C., Shen, L., & Qian, G. (2023). Online Hybrid Neural Network for Stock Price Prediction: A Case Study of High-Frequency Stock Trading in the Chinese Market. *Econometrics*, *11*(2), 13.
[58] Norton, V. (2011). Adjusted closing prices. *arXiv preprint arXiv:1105.2956*.
[59] Hernandez-Leal, P., Kartal, B., & Taylor, M. E. (2019, October). Agent modeling as auxiliary task for deep reinforcement learning. In *Proceedings of the AAAI conference on artificial intelligence and interactive digital entertainment* (Vol. 15, No. 1, pp. 31-37).
[60] Wang, R., Wei, H., An, B., Feng, Z., & Yao, J. (2020). Deep stock trading: A hierarchical reinforcement learning framework for portfolio optimization and order execution. *arxiv preprint arxiv:2012.12620*.
[61] Lo, A. W. (2004). The adaptive markets hypothesis: Market efficiency from an evolutionary perspective. *Journal of Portfolio Management, Forthcoming*.
[62] Duan, Y., Wang, L., Zhang, Q., & Li, J. (2022, June). Factorvae: A probabilistic dynamic factor model based on variational autoencoder for predicting cross-sectional stock returns. In *Proceedings of the AAAI Conference on Artificial Intelligence* (Vol. 36, No. 4, pp. 4468-4476).
[63] Liu, X. Y., Yang, H., Gao, J., & Wang, C. D. (2021, November). FinRL: Deep reinforcement learning framework to automate trading in quantitative finance. In *Proceedings of the second ACM international conference on AI in finance* (pp. 1-9).
[64] Cover, T. M. (1991). Universal portfolios. *Mathematical finance*, *1*(1), 1-29.
[65] Borodin, A., El-Yaniv, R., & Gogan, V. (2000). On the competitive theory and practice of portfolio selection. In *LATIN 2000: Theoretical Informatics: 4th Latin American Symposium, Punta del Este, Uruguay, April 10-14, 2000 Proceedings 4* (pp. 173-196). Springer Berlin Heidelberg.
[66] Li, B., & Hoi, S. C. (2014). Online portfolio selection: A survey. *ACM Computing Surveys (CSUR)*, *46*(3), 1-36
[67] Cover, T. M., & Ordentlich, E. (1996). Universal portfolios with side information. *IEEE Transactions on Information Theory*, *42*(2), 348-363.
[68] Helmbold, D. P., Schapire, R. E., Singer, Y., & Warmuth, M. K. (1998). On-line portfolio selection using multiplicative updates. *Mathematical Finance*, *8*(4), 325-347.
[69] Borodin, A., El-Yaniv, R., & Gogan, V. (2003). Can we learn to beat the best stock. *Advances in Neural Information Processing Systems*, *16*.
[70] Li, B., Zhao, P., Hoi, S. C., & Gopalkrishnan, V. (2012). PAMR: Passive aggressive mean reversion strategy for portfolio selection. *Machine learning*, *87*, 221-258.
[71] Li, B., Hoi, S. C., Zhao, P., & Gopalkrishnan, V. (2011, June). Confidence weighted mean reversion strategy for on-line portfolio selection. In *Proceedings of the Fourteenth International Conference on Artificial Intelligence and Statistics* (pp. 434-442). JMLR Workshop and Conference Proceedings.
[72] Li, B., & Hoi, S. C. (2012). On-line portfolio selection with moving average reversion. *arXiv preprint arXiv:1206.4626*.
[73] Huang, D., Zhou, J., Li, B., HOI, S., & Zhou, S. (2012). Robust Median Reversion Strategy for On-Line Portfolio Selection.(2013). In *Proceedings of the Twenty-Third International Joint Conference on Artificial Intelligence: IJCAI 2013*.
[74] Gao, L., & Zhang, W. (2013, August). Weighted moving average passive aggressive algorithm for online portfolio selection. In *2013 5th International Conference on Intelligent Human-Machine Systems and Cybernetics* (Vol. 1, pp. 327-330). IEEE.
[75] Györfi, L., Lugosi, G., & Udina, F. (2006). Nonparametric kernel-based sequential investment strategies. *Mathematical Finance: An International Journal of Mathematics, Statistics and Financial*

*Economics*, *16*(2), 337-357.

[76] Li, B., Hoi, S. C., & Gopalkrishnan, V. (2011). Corn: Correlation-driven nonparametric learning approach for portfolio selection. *ACM Transactions on Intelligent Systems and Technology (TIST)*, *2*(3), 1-29.

[77] Agarwal, A., Hazan, E., Kale, S., &Schapire, R. E. (2006). Algorithms for portfolio management based on the Newton method. In *ACM: Proceedings of the 23rd international conference on machine learning (pp. 9–16)*. http://dx.doi.org/10.1145/ 1143844.1143846.

[78] Jorion, P. (1986). Bayes-Stein estimation for portfolio analysis. *Journal of Financial and Quantitative analysis*, *21*(3), 279-292.

[79] Kan, R., & Zhou, G. (2007). Optimal portfolio choice with parameter uncertainty. *Journal of Financial and Quantitative Analysis*, *42*(3), 621-656.

[80] Zeng, A., Chen, M., Zhang, L., & Xu, Q. (2023, June). Are transformers effective for time series forecasting?. In *Proceedings of the AAAI conference on artificial intelligence* (Vol. 37, No. 9, pp. 11121-11128).

[81] Wu, H., Xu, J., Wang, J., & Long, M. (2021). Autoformer: Decomposition transformers with auto-correlation for long-term series forecasting. *Advances in neural information processing systems*, *34*, 22419-22430.

[82] Zhou, H., Zhang, S., Peng, J., Zhang, S., Li, J., Xiong, H., & Zhang, W. (2021, May). Informer: Beyond efficient transformer for long sequence time-series forecasting. In *Proceedings of the AAAI conference on artificial intelligence* (Vol. 35, No. 12, pp. 11106-11115).

[83] Nie, Y., Nguyen, N. H., Sinthong, P., & Kalagnanam, J. (2022). A time series is worth 64 words: Long-term forecasting with transformers. *arXiv preprint arXiv:2211.14730*.

[84] Xiao, Y., & Valdez, E. A. (2015). A Black–Litterman asset allocation model under Elliptical distributions. *Quantitative Finance*, *15*(3), 509-519.

[85] Sutton, R. S. (1988). Learning to predict by the methods of temporal differences. *Machine learning*, *3*, 9-44.

# Appendix A

The list of the Dow Jones Industrial Average (DJIA) components for portfolio construction and their respective tickers, names, and categories

| No. | Ticker | Name | Category |
|---|---|---|---|
| 1 | MMM | 3M Company | Industrial |
| 2 | AXP | American Express Company | Financial |
| 3 | AMGN | Amgen Inc. | Healthcare |
| 4 | AAPL | Apple Inc. | Technology |
| 5 | BA | Boeing Company | Industrial |
| 6 | CAT | Caterpillar Inc. | Industrial |
| 7 | CVX | Chevron Corporation | Energy |
| 8 | CSCO | Cisco Systems, Inc. | Technology |
| 9 | KO | Coca-Cola Company | Consumer goods |
| 10 | GS | Goldman Sachs Group, Inc. | Financial |
| 11 | HD | The Home Depot, Inc. | Consumer services |
| 12 | HON | Honeywell International Inc. | Industrial |
| 13 | IBM | IBM Corporation | Technology |
| 14 | INTC | Intel Corporation | Technology |
| 15 | JNJ | Johnson & Johnson | Healthcare |
| 16 | JPM | JPMorgan Chase & Co. | Financial |
| 17 | MCD | McDonald's Corporation | Consumer services |
| 18 | MRK | Merck & Co., Inc. | Healthcare |
| 19 | MSFT | Microsoft Corporation | Technology |
| 20 | NKE | Nike, Inc. | Consumer goods |
| 21 | PG | Procter & Gamble Company | Consumer goods |
| 22 | CRM | Salesforce.com, Inc. | Technology |
| 23 | TRV | The Travelers Companies, Inc. | Financial |
| 24 | UNH | UnitedHealth Group Incorporated | Healthcare |
| 25 | VZ | Verizon Communications Inc. | Communication |
| 26 | V | Visa Inc. | Financial |
| 27 | WBA | Walgreens Boots Alliance, Inc. | Consumer goods |
| 28 | WMT | Walmart Inc. | Consumer services |
| 29 | DIS | The Walt Disney Company | Consumer services |

# Appendix B

Hyper-parameters in the paper

| Where | Hyper-parameters | Value |
|---|---|---|
| Hyper-parameters in the state tensor | Number of trading days $K$ included in a trading period | 5 |
| Hyper-parameters in the state tensor | Number of trading periods $M$ included in the historical return tensor | 40 |
| Hyper-parameters in the portfolio | Number of stocks $n$ included in the historical return tensor | 29 |
| Hyper-parameters in the portfolio | The annual borrowing rate $r_l$ | 3% |
| Hyper-parameters in the portfolio | The annual borrowing rate $r_s$ for stocks | 3% |
| Hyper-parameters in the reward function | The parameters $\lambda_1$ to describe the risk aversion in the reward function | 10 |
| Hyper-parameters in the reward function | The parameters $\lambda_2$ to limit the transaction scale in the reward function | 0.001 |
| Hyper-parameters in the reward function | The parameters $\lambda_3$ to describe the risk aversion in the target value calculation | 50 |
| Hyper-parameters in the DRL training | Target step $M$ | 1080 |
| Hyper-parameters in the DRL training | Minibatch size $N$ | 128 |
| Hyper-parameters in the DRL training | Size of the Replay buffer $B_{au}$ for the auxiliary agent | 2^14 |
| Hyper-parameters in the DRL training | Size of the Replay buffer $B_{ex}$ for the executive agent | 2^14 |
| Hyper-parameters in the DRL training | Total step $S_{au}^{total}$ for the training of the auxiliary agent | 3e5 |
| Hyper-parameters in the DRL training | Total step $S_{ex}^{total}$ for the training of the executive agent | 6e5 |
| Hyper-parameters in the DRL training | Learning rates $\alpha_1$ for the auxiliary agent's policy network $\pi_\phi^{au}(S_t)$ | 1e-5 |
| Hyper-parameters in the DRL training | Learning rates $\alpha_2$ for the executive agent's policy network $\pi_\theta^{ex}(S_t)$ | 1e-6 |
| Hyper-parameters in the DRL training | Learning rates $\alpha_3$ for the critic $Q_\omega(s_t, a_t)$, | 1e-8 |

# Appendix C

| Algorithm 1. The training process of the DRL agent |
|---|

Input:
policy function $\pi_\phi^{au}(s_t)$ for the auxiliary agent,
policy function $\pi_\theta^{ex}(s_t)$ for the executive agent,
critic network $Q_\omega(s_t, a_t)$ for the executive agent,
learning rates $\alpha_1$ for the auxiliary agent's policy network $\pi_\phi^{au}(s_t)$,
learning rates $\alpha_2$ for the executive agent's policy network $\pi_\theta^{ex}(s_t)$,
learning rates $\alpha_3$ for the executive agent's critic network $Q_\omega(s_t, a_t)$,
minibatch size of the auxiliary agent $N^{au}$,
minibatch size of the executive agent $N^{ex}$,
target step $M^{au}$ of the auxiliary agent,
target step $M^{ex}$ of the executive agent,
discount factor $\gamma$,
momentum coefficient $\tau$,
number of trading periods $T_f^{(tr)}$ in the training set,
total step $S_{au}^{total}$ for the training of the auxiliary agent,
total step $S_{ex}^{total}$ for the training of the executive agent.

---

(Begin to train the auxiliary agent)
1. Initialize policy network $\pi_\phi^{au}(s_t)$ with parameters $\phi$.
2. Initialize the accumulated steps $S_{au} = 0$ for training the auxiliary agent, trading period $t = 1$, current total assets $v_0 = T_0 = c_0 = 1e8$, $train_{au}$ = True.
3. Build the replay buffer $B^{au}$ for the auxiliary agent.
4. Observe the current state $s_1^{au}$.

(Update the replay buffer $B^{au}$)
5. If $train_{au}$ do:
6.     For $n = 1, 2, \ldots, M^{au}$ do
7.         Select auxiliary action $a_t^{au}$ based on the policy function $\pi_\phi^{au}$ and the current state $s_t^{au}$:
$$a_t^{au} = \pi_\phi^{au}(s_t^{au}).$$
8.         Enter into the next trading period.
9.         Observe the state $s_{t+1}^{au}$ from the environment.
10.        Store the tuple $\{s_t^{au}, a_t^{au}, s_{t+1}^{au}\}$ into the replay buffer $B^{au}$.
11.        Let $t = t + 1$.
12.        If $t > T_f^{(tr)}$ do:
13.            Reset the trading period: $t = 1$.
14.            Observe the current state $s_1^{au}$

(Train the policy network $\pi_\phi^{au}$ of the auxiliary agent)
15.     Calculate the length of the rely buffer $L^{au}$.
16.     For $n = 1, 2, \ldots, \lfloor L^{au}/N^{au} \rfloor$ do:
17.         Randomly sample a mini-batch of the states $\{s_i^{au}, a_i^{au}, s_{i+1}^{au}\}_{i=1}^{N^{au}}$ from the replay buffer $B^{au}$.
18.         Update the parameters of the policy function $\phi$:
$$\phi = \phi + \alpha_1 \frac{1}{N^{au}} \sum_{i=1}^{N^{au}} \nabla_\phi r^{au}(\pi_\phi^{au}(s_i^{au}), s_i^{au} | \lambda_1, \lambda_2, \lambda_3)$$

(Calculate the total step and decide whether to finish the training of the auxiliary agent)
19.         Update the accumulated steps:
$$S_{au} = S_{au} + M^{au}.$$
20.     If $S_{au} > S_{au}^{total}$ do:
           $train_{au}$ = False

(The training of the auxiliary agent is finished, and then we begin to train the executive agent)
21. Initialize policy function (actor network) $\pi_\theta^{ex}(s_t)$ with parameters $\theta$.
22. Initialize critic network $Q_\omega(s_t, a_t)$ with parameters $\omega$.
23. Initialize the target policy network (target actor network) $\pi_{\theta'}(s_t)$ and target critic network $Q_{\omega'}(s_t, a_t)$.
24. Initialize the Accumulated steps $S_e = 0$ for training the executive agent, trading period $t = 1$, current total assets $v_0 = T_0 = c_0 = 1e8$, $train_{ex} = $ True.
25. Build the replay buffer $B^{ex}$ for the executive agent.
26.
27. Construct random process $\mathcal{N}$ for action exploration.
28. Observe the initial state $s_1^{au}$ for the auxiliary agent from the environment at the beginning of the first trading period.
29. Let the auxiliary agent determine the baseline portfolio weights $w_1^{au}$ for the executive agent.
30. Describe the current state $s_1^{ex}$ for the executive agent based on the historical return tensor $X_1$ and the baseline portfolio weights $w_1^{au}$ determined by the auxiliary agent at the beginning of the first trading period:
$$s_1^{ex} = \langle w_1^{au}, X_1 \rangle.$$

(Update the replay buffer $B^{ex}$)
31. If $train_{ex}$ do:
32.     For $n = 1, 2, \ldots, M^{ex}$ do
33.         Select executive action $a_t^{ex}$ based on the policy function and the current state $s_t^{ex}$:
$$a_t^{ex} = \pi_\phi^{ex}(s_t^{ex}) + \mathcal{N},$$
34.         Execute the action $a_t^{ex}$ and observe the reward $r_t^{ex}$ from the environment.
35.         Enter into the next trading period.
36.         Observe the state $s_{t+1}^{au}$ for the auxiliary agent.
37.         Let the auxiliary agent determine the baseline portfolio weights $w_{t+1}^{au}$ based on the policy function $\pi_\phi^{au}$ and the current state $s_{t+1}^{au}$:
$$a_{t+1}^{au} = \pi_\phi^{au}(s_{t+1}^{au}).$$
38.         Describe the state $s_{t+1}^{ex}$ for the executive agent based on the historical return tensor $X_{t+1}$ and the baseline portfolio weights $w_{t+1}^{au}$:
$$s_{t+1}^{ex} = \langle w_{t+1}^{au}, X_{t+1} \rangle.$$
39.         Store the tuple $\{s_t^{ex}, a_t^{ex}, r_t^{ex}, s_{t+1}^{ex}\}$ into the replay buffer $B^{ex}$.
40.         Let $t = t + 1$.
41.         If $t > T_f^{(tr)}$ do:
42.             Reset the trading period: $t = 1$.
43.             Observe the initial state $s_1^{au}$ for the auxiliary agent from the environment at the beginning of the first trading period.
44.             Let the auxiliary agent determine the baseline portfolio weights $w_1^{au}$ for the executive agent.
45.             Describe the current state $s_1^{ex}$ for the executive agent based on the historical return tensor $X_1$ and the baseline portfolio weights $w_1^{au}$ at the beginning of the first trading period:
$$s_1^{ex} = \langle w_1^{au}, X_1 \rangle.$$
46.

(Train the policy network $\pi_\theta^{ex}$ of the executed agent)
47.     Calculate the length of the rely buffer $L^{ex}$.
48.     For $n = 1, 2, \ldots, \lfloor L^{ex}/N^{ex} \rfloor$ do:
49.         Randomly sample a mini-batch of the states $\{s_i^{ex}, a_i^{ex}, r_i^{ex}, s_{i+1}^{ex}\}_{i=1}^{N^{ex}}$ from the replay buffer $B^{ex}$.
50.         Update the parameters of the critic network $\omega$:
$$\omega = \omega - \alpha_3 \frac{1}{N^{ex}} \sum_{i=1}^{N^{ex}} \nabla_\omega [Q_\omega(s_i^{ex}, a_i^{ex}) - (r_i^{ex} + \gamma\, Q_{\omega'}(s_{i+1}^{ex}, \pi_{\theta'}^{ex}(s_{i+1}^{ex})))]^2$$

| | |
|---|---|
| 51. | Soft update the parameters of the target critic $\omega'$:<br>$$\omega' \leftarrow \tau\omega + (1-\tau)\omega'$$ |
| 52. | Update the parameters of the policy function $\theta$:<br>$$\theta = \theta + \alpha_2 \frac{1}{N^{ex}} \sum_i^{N^{ex}} \nabla_{a_i^e} Q(s_i^{ex}, \pi_\theta^{ex}(s_i^{ex})) \nabla_\theta \pi_\theta^{ex}(s_i^{ex})$$ |
| 53. | Soft update the parameters of the target policy function $\theta'$:<br>$$\theta' \leftarrow \tau\theta + (1-\tau)\theta'$$ |
| 54. | Update the accumulated steps:<br>$$S_{ex} = S_{ex} + M^{ex}.$$ |
| 55. | If $S_{ex} > S_{ex}^{total}$ do:<br>$train_{ex}$ = False |

# Appendix D

Similar to the numerical changes trajectories of the significant indices in Section 4.4, we also track the numerical changes trajectories of these indices in experiments 2-4. The empirical results are given in the following figures.

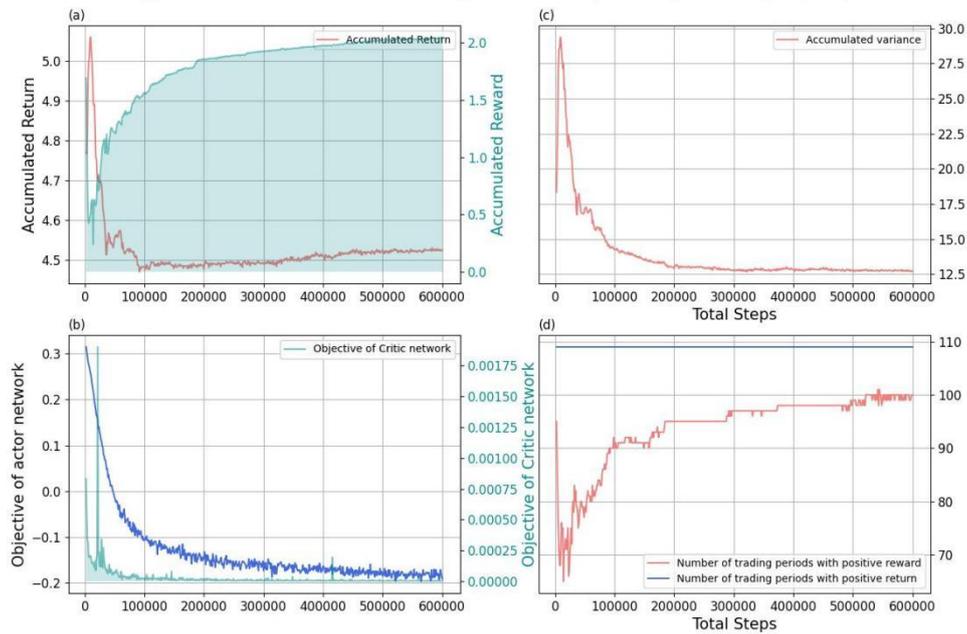

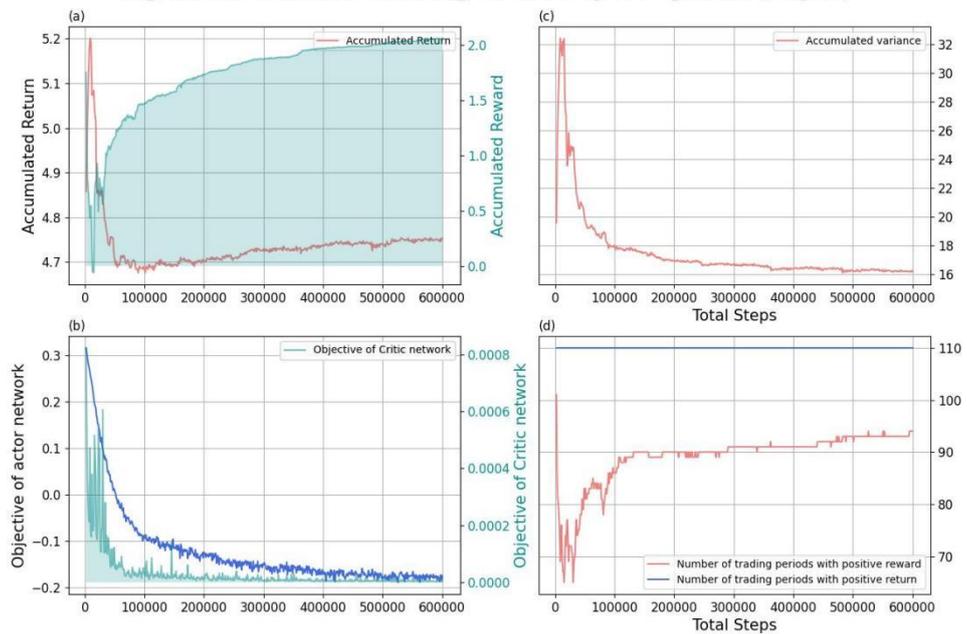

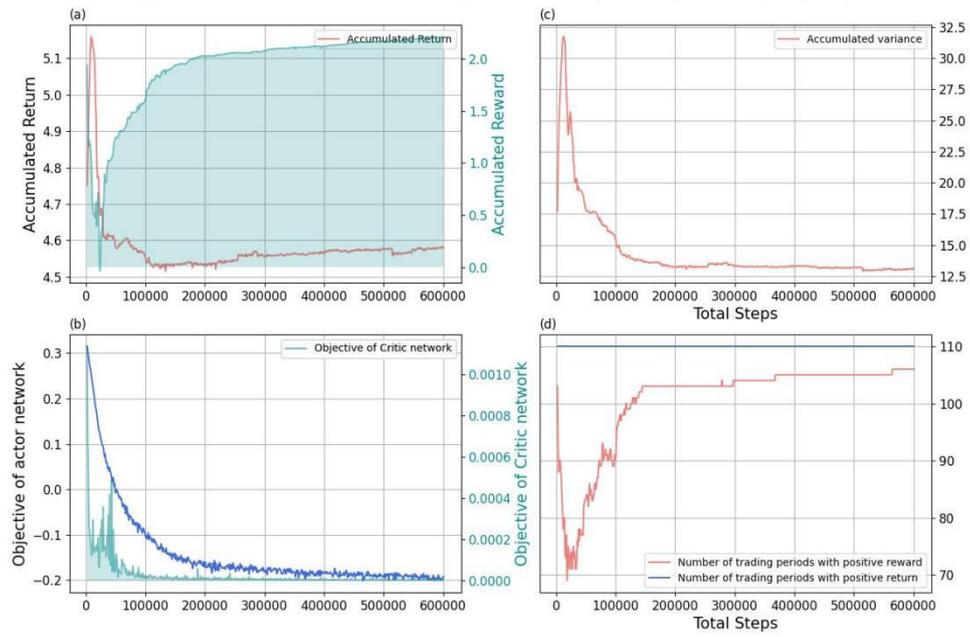

Significant Indices Tracking of the Expert System (Exp.4)